\documentclass[10pt,twocolumn,letterpaper]{article}

\usepackage{iccv}
\usepackage{times}
\usepackage{epsfig}
\usepackage{graphicx}
\usepackage{amsmath}
\usepackage{amssymb}
\usepackage{booktabs}
\usepackage[table,xcdraw]{xcolor}
\usepackage{colortbl}
\usepackage{upgreek}
\usepackage{color}
\usepackage{threeparttable}

\usepackage[accsupp]{axessibility} 

\usepackage[pagebackref=true,breaklinks=true,letterpaper=true,colorlinks,bookmarks=false]{hyperref}

\usepackage[capitalize]{cleveref}
\crefname{section}{Sec.}{Secs.}
\Crefname{section}{Section}{Sections}
\Crefname{table}{Table}{Tables}
\crefname{table}{Tab.}{Tabs.}

\iccvfinalcopy 

\def\iccvPaperID{6967} 

\ificcvfinal\fi

\begin{document}

\title{3D Implicit Transporter for Temporally Consistent Keypoint Discovery}

\author{
Chengliang Zhong\textsuperscript{1,2},
Yuhang Zheng\textsuperscript{3},
Yupeng Zheng\textsuperscript{4},
Hao Zhao\textsuperscript{2}\thanks{Corresponding author},
Li Yi\textsuperscript{5},
Xiaodong Mu\textsuperscript{1},\\
Ling Wang\textsuperscript{1}, 
Pengfei Li\textsuperscript{2},
Guyue Zhou\textsuperscript{2},
Chao Yang\textsuperscript{6},
Xinliang Zhang\textsuperscript{2},
Jian Zhao\textsuperscript{7,8,9}\\
\textsuperscript{1}Xi'an Research Institute of High-Tech,
\textsuperscript{2}AIR, Tsinghua University, 
\textsuperscript{3}Beihang University,\\
\textsuperscript{4}Institute of Automation, Chinese Academy of Sciences,
\textsuperscript{5}IIIS, Tsinghua University,\\
\textsuperscript{6}Shanghai AI Laboratory,
\textsuperscript{7}Institute of North Electronic Equipment,\\
\textsuperscript{8}Intelligent Game and Decision Laboratory,
\textsuperscript{9}Peng Cheng Laboratory\\
{\tt\small zhongcl19@mails.tsinghua.edu.cn, zhaohao@air.tsinghua.edu.cn}\\
}

\maketitle
\ificcvfinal\thispagestyle{empty}\fi

\begin{abstract}
 Keypoint-based representation has proven advantageous in various visual and robotic tasks. However, the existing 2D and 3D methods for detecting keypoints mainly rely on geometric consistency to achieve spatial alignment, neglecting temporal consistency. To address this issue, the Transporter method was introduced for 2D data, which reconstructs the target frame from the source frame to incorporate both spatial and temporal information. However, the direct application of the Transporter to 3D point clouds is infeasible due to their structural differences from 2D images. Thus, we propose the first 3D version of the Transporter, which leverages hybrid 3D representation, cross attention, and implicit reconstruction. We apply this new learning system on 3D articulated objects and nonrigid animals (humans and rodents) and show that learned keypoints are spatio-temporally consistent. Additionally, we propose a closed-loop control strategy that utilizes the learned keypoints for 3D object manipulation and demonstrate its superior performance. Codes are available at \href{https://github.com/zhongcl-thu/3D-Implicit-Transporter}{https://github.com/zhongcl-thu/3D-Implicit-Transporter}.
\end{abstract}

\section{Introduction}

The ability to establish correspondences in temporal inputs is a hallmark of the human visual system, and this ability has been verified by developmental biologists \cite{spelke1990principles} as the enabling factor of object perception. Specifically, infants can naturally separate different objects by considering pixels that move together. Meanwhile, establishing dense correspondences from image sequences (i.e., optical flow) \cite{baker2011database,barron1994performance,butler2012naturalistic,dosovitskiy2015flownet,sun2010secrets,sun2018pwc,sun2022disentangling} is also one of the oldest computer vision topics, dating back to the birth of this discipline \cite{horn1981determining}.


On the other hand, keypoints are preferred as a compact mid-level representation in many scenarios, like visual recognition \cite{sivic2003video}, pose estimation\cite{zhao2020fine}, reconstruction \cite{mur2015orb}, and robotic manipulation \cite{qin2020keto}.
Sparse correspondence from keypoints \cite{lowe2004distinctive, bay2008speeded, calonder2010brief, rublee2011orb} is another fundamental vision topic. Most of the 2D and 3D keypoint detection methods \cite{zhong2022sim2real,li2019usip} depend on the consistency under geometric transformations to achieve spatial alignment of keypoints. However, these methods are limited in their ability to identify temporally consistent keypoints, which is crucial for representing movable or deformable objects such as the human body whose shape and topology may vary over time.
So does there exist a generic principle that reflects how humans extract spatiotemporally consistent keypoints? One (possible) such principle is that good mid-level representation can be used to re-synthesize raw visual inputs. This principle has been explored in legacy methods like FRAME \cite{zhu1998filters}, but limited by the modeling power of generative models at that age, and this principle has not seen much success.

\begin{figure}
  \includegraphics[width=0.99\linewidth]{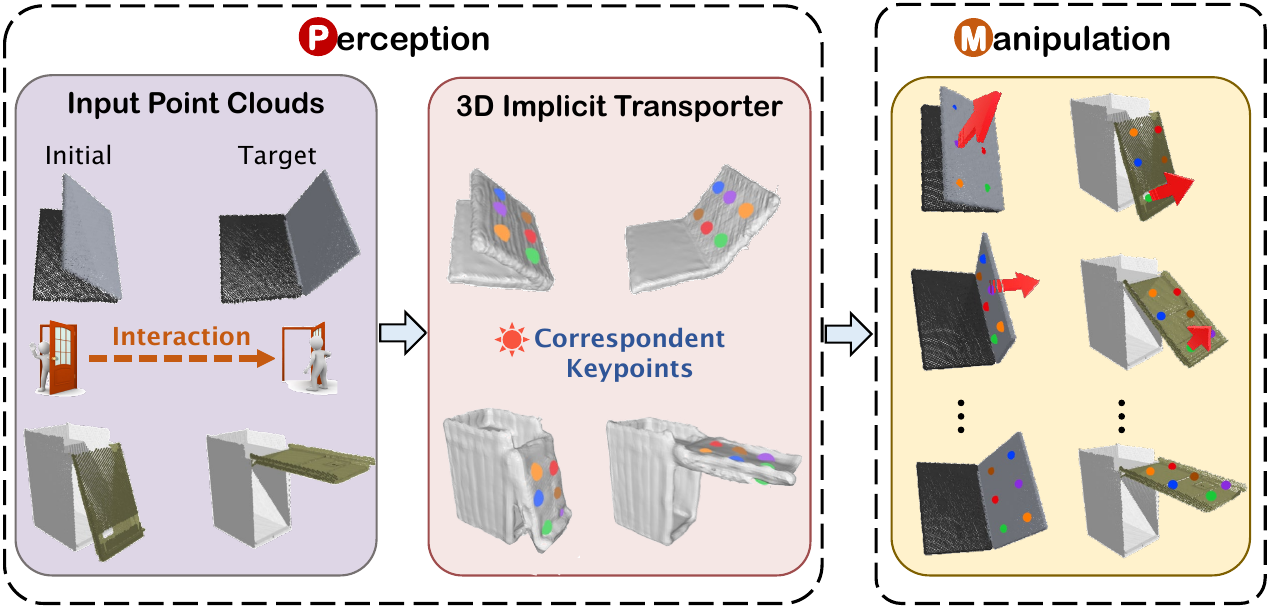}
    \caption{Given paired point clouds, our 3D Implicit Transporter leverages the object/part motion to discover temporally consistent keypoints and recover the underlying shape for each input. Moreover, the learned keypoints can serve downstream robotics tasks, such as articulated object manipulation.}
    \label{fig:teaser}
    \vskip -0.2in
\end{figure}

Recently, a method named Transporter \cite{kulkarni2019unsupervised} has been proposed in the 2D domain, successfully connecting keypoint extraction and correspondence establishment in a self-supervised manner, using exactly the aforementioned principle. Thanks to the strong power of modern image reconstruction networks, this method can extract meaningful keypoints from image sequences without using any human annotation. Transporter is both a useful tool and an elegant formulation that (potentially) mimics how the human visual system extracts keypoints. However, to the best of our knowledge, this methodology has not been translated into the 3D domain. 
Because the process of 2D feature transportation is implemented on regular data formats, such as 2D image grids, it is not applicable to point clouds that allow for non-uniform point spacing.
Another reason we believe is that 3D reconstruction from keypoints is a more challenging setting.

As such, we propose the first 3D Transporter in the literature, based upon three core components: hybrid 3D representation architectures for 3D feature transportation, cross-attention 
for better keypoint discovery, 
and an implicit geometry decoder for 3D reconstruction. Our method takes as input two point clouds containing moving objects or object parts (Fig.~\ref{fig:teaser} left panel).
Then by watching these two states solely, our 3D Implicit Transporter extracts temporally consistent keypoints and a surface occupancy field for each state, in a self-supervised manner (Fig.~\ref{fig:teaser} middle panel). The method reconstructs the shape of the target state by transporting explicit feature grids from the initial state according to the locations of detected keypoints. 
Via extensive evaluations on the PartNet-Mobility dataset \cite{xiang2020sapien} and ITOP dataset \cite{haque2016towards}, we demonstrate significantly improved perception performance in terms of spatiotemporal consistency of keypoints over state-of-the-art counterparts. The qualitative results on the Rodent3D\cite{patel2023animal} dataset also show our keypoints 
consistently capture the rodent’s skeletal structure.

Besides, we also explore how well our self-supervised mid-level representation (3D keypoints) can serve downstream robotic applications (Fig.~\ref{fig:teaser} right panel). We choose articulated object manipulation, which requires sophisticated 3D reasoning about the kinematic structure and part movement, as the benchmark. Existing methods often rely on object-agnostic affordance-based representation \cite{xu2022umpnet,mo2021where2act,wang2022adaafford}. Our 3D keypoint representation is also object-agnostic, but we demonstrate that our approach offers two distinct advantages over these methods: 1) the efficient learning formulation of ours does not involve costly trial-and-error interaction in simulators; 
2) we leverage spatio-temporally aligned 3D keypoints to provide a structured understanding of objects, enabling the design of an effective closed-loop manipulation strategy.

To summarize, we have the following contributions:

\begin{itemize}
\item We propose the first 3D Implicit Transporter formulation that extracts 3D correspondent keypoints from temporal point cloud inputs, using 3D feature grid transportation, attentional keypoint detection, and target shape reconstruction.
\item Based on the extracted 3D keypoint representation, we build a closed-loop manipulation strategy and demonstrate it successfully addresses the manipulation of many articulated objects in an object-agnostic setting.
\item We extensively benchmark the perception and manipulation performance of 3D Implicit Transporter and report state-of-the-art results on public benchmarks. 

\end{itemize}


\section{Related Work}
\subsection{3D Keypoint Detection}
Detecting 3D keypoints from point clouds has drawn a lot of attention in vision and robotics \cite{rahmani2014hopc, suwajanakorn2018discovery, bai2020d3feat, li2019usip, Zeng20173DMatchLL, zhou2016fast}. 
Traditional hand-crafted methods predict salient points according to local geometric statistics of inputs, such as density \cite{zhong2009intrinsic} and curvature \cite{lee2005mesh}. Modern learning-based approaches employ the consistency of keypoint coordinates \cite{li2019usip} or saliency scores \cite{zhong2022snake} under rigid transformations to formulate keypoint detection as a self-supervised task. 
However, they are unable to ensure temporally consistent keypoint detection of non-rigid objects whose shapes are significantly changed after movement. 
To achieve that, recent research has explored discovering temporally aligned keypoints from given image videos. Most of them \cite{minderer2019unsupervised, sun2022self, gopalakrishnan2020unsupervised, chen2021unsupervised, kulkarni2019unsupervised, wang2021unsupervised} consider the keypoint learning problem as a signal reconstruction process. 
For example, Minderer \textit{et al.} \cite{minderer2019unsupervised} and Jennifer \cite{sun2022self}  proposed to reconstruct a future frame by using features of the current frame and future keypoints. 
Despite their better results, these methods all concentrate on 2D keypoint discovery. As far as we know, there is seldom work investigating the task of 3D temporally consistent keypoint detection. 

\subsection{Neural Implicit Representation}
Recently, multiple works \cite{liu2020neural, mildenhall2020nerf, schwarz2020graf, Li2023LODELC, park2019deepsdf, sitzmann2020implicit} have focused on implicit geometric representation. It intends to parameterize a signal as a continuous function by a neural network that could decode complex shape topologies of discrete inputs \cite{park2019deepsdf, convonet} It shows great achievements of the implicit neural function on the grasp pose generation \cite{jiang2021synergies}, articulated model estimation \cite{jiang2022ditto, mu2021sdf}, and object pose representation \cite{simeonov2022neural}. More recently, it has been proved that entangling an implicit shape decoder \cite{zhong2022snake} instead of a coordinate decoder \cite{you2022ukpgan} encourages the model to predict more semantically consistent keypoints. Inspired by these works, we exploit the implicit occupancy function to reconstruct the underlying shape of the transported 3D objects. 

\subsection{Perception and Manipulation of Articulated Objects}
Previous research has explored various techniques to understand and represent articulated objects, such as kinematic graphs \cite{katz2008manipulating, abdul2022learning}, 6D pose estimation \cite{li2020category, weng2021captra}, part segmentation \cite{wang2019shape2motion, xiang2020sapien, yi2018deep}, deformation flow \cite{yi2018deep}, joint parameters \cite{jiang2022ditto, nie2022structure} and so on. However, most of them require ground truth knowledge of the object or are category-dependent. Unlike them, we use sparse correspondence from keypoints to capture the critical part of articulated objects without human labels, which can also generalize to unseen categories. Albeit the fruitful progress in vision, one could not directly infer an action from perceptual outputs to manipulate articulated objects. Therefore, recent works have proposed manipulation-centric visual representations, like
visual affordance \cite{mo2021where2act,xu2022umpnet,wu2021vat} or dense articulation flow \cite{EisnerZhang2022FLOW}. 
However, they need substantial trial-and-error interactions in simulation or ground-truth geometric knowledge. In contrast, the keypoint learning of ours is unsupervised and effective, and the correspondences between keypoints can also serve robotic manipulation well.

\begin{figure*}[t]
  \centering
  \includegraphics[width=0.95\linewidth]{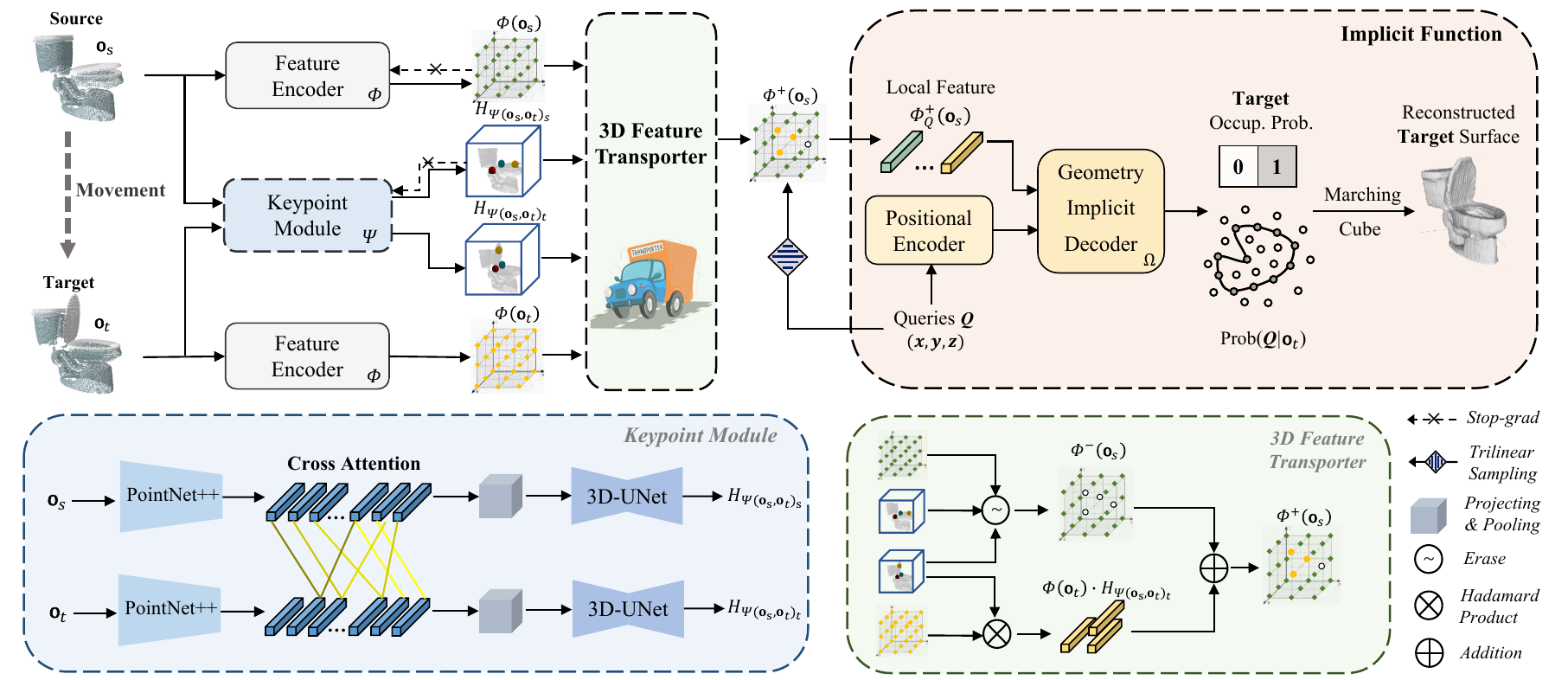}
  \caption{\textbf{Architecture of our 3D Implicit Transporter.} 
  The network consists of (1) a feature encoder $\Phi$ that extracts features for transportation; (2) a keypoint module $\Psi$ to indicate where to transport; (3) a 3D feature transporter that reconstruct the feature volume of interest; and (4) a geometry implicit decoder that allows self-supervision using solely input point clouds. The overall process involves extracting $m$ corresponding 3D keypoints from two frames and using them to transport the features of the target frame to the source frame based on their keypoint locations. The transported features are then fed into the implicit decoder to reconstruct the target shape.}
  \label{fig:main}
  \vspace{-3mm}
\end{figure*}

\section{Method}

Our perception method presents a new formulation to discover temporally and spatially consistent 3D keypoints of a moving object or object part in a sequence of point clouds in a self-supervised manner.
Once trained, the learned keypoints are used to devise a strategy for the manipulation of an articulated object from its starting state to a goal state, thus avoiding the costly trial-and-error interaction typically utilized in \cite{xu2022umpnet}.


\subsection{Temporal 3D Keypoint Discovery}
In accordance with the formulation in \cite{kulkarni2019unsupervised}, we consider a dataset comprising pairs of frames extracted from a series of trajectories, where each frame is represented as a 3D point cloud instead of an image. Two frames in a pair are distinguished by differences solely in the pose/geometry of the objects. Our goal is to find the corresponding keypoints describing the object or object part motion from the source to the target frame. We tackle this problem by reconstructing the underlying shape of the target frame from the source frame. Fig. \ref{fig:main} summarizes our method, and the subsequent subsections provide further details on its components.


\vspace{-2mm}
\subsubsection{3D Feature Transporter}

\textbf{Hybrid 3D Representation.}
As per the formulation in 2D Transporter, feature transportation occurs between uniform data, such as 2D images, which is not feasible for point clouds that are in an irregular format. A straight way is to convert point clouds to uniform 3D voxel grids before inputting them to a neural network. However, converting raw point clouds into voxels inevitably introduces quantization errors that break the intrinsic geometric patterns (e.g., isometry) of 3D data. 
Although a high-resolution volumetric representation could compensate this information loss, both computational cost and memory requirements escalate cubically with voxel resolution. On the contrary, point-based models \cite{fan2017point, qi2017pointnetplusplus} lead to a significant reduction in memory usage due to the sparse representation.
As such, we utilize the point-based backbone to extract local features from sparse points and then leverage the voxel-based model for transporting local features.

Given a frame $\boldsymbol{{\rm o}} \in \mathbb{R}^{N_1 \times 3}$, where $N_1$ is the number of input points, we exploit a PointNet \cite{qi2017pointnetplusplus} $P$ to get point features $P(\boldsymbol{{\rm o}}) \in \mathbb{R}^{N_1 \times C_1}$, where $C_1$ is the dimension of features.
These features are then locally pooled and projected into structured volumes $\boldsymbol{{\rm v}} \in \mathbb{R}^{C_2\times C_h \times C_w \times C_d}$, where $C_h$, $C_w$ and $C_d$ are the number of voxels in three orthogonal axes. Afterwards, the feature volumes are processed with a 3D UNet \cite{cciccek20163d} $U$, resulting in the outputs $U(\boldsymbol{{\rm v}}) \in \mathbb{R}^{C_3 \times C_h \times C_w \times C_d}$. The above point and voxel-based models are denoted as the feature encoder $\mit \Phi$ in Fig.~\ref{fig:main}. 

\textbf{Attentional Keypoint Detection.}
When we are asked to find moving objects or object parts between paired frames, we adopt an iterative process whereby we inspect and sift through multiple tentative regions in both frames. 
However, extracting keypoints on mobile parts using a single frame, as done by 2D Transporter\footnote{The reconstruction objective requires two frames.}, can be inherently ambiguous, particularly when multiple potential mobile parts exist in each frame. Therefore, inspired by \cite{superglue,jiang2022ditto,li2022lepard,zeng2021corrnet3d}, we propose to use a cross-attention module to aggregate geometric features from both frames to locate keypoints.

Specifically, we utilize a point-based model (not shared with $\mit \Phi$) to extract multi-level features for the input point clouds and correlate paired inputs at a coarse level to reduce computational costs, as done in \cite{jiang2022ditto}.
Given a frame pair $\boldsymbol{{\rm o}}_s, \boldsymbol{{\rm o}}_t$, we exploit a shared PointNet++ \cite{qi2017pointnetplusplus} $\hat{P}$ to get two down-sampled point features $\boldsymbol{{\rm f}}_s=\hat{P}(\boldsymbol{{\rm o}}_s)$ and $\boldsymbol{{\rm f}}_{t}=\hat{P}(\boldsymbol{{\rm o}}_t)$, where $\boldsymbol{{\rm f}}_s, \boldsymbol{{\rm f}}_{t} \in \mathbb{R}^{N_2 \times C_4}$. Then, a cross attention block \cite{vaswani2017attention} is used to mix point features of paired inputs, achieved by:
\vspace{-5mm}
\begin{align}
\label{equ:atten1}
&\boldsymbol{{\rm z}}_s = {\rm softmax}(\frac{\boldsymbol{{\rm f}}_s\boldsymbol{{\rm f}}_t^T}{\sqrt{C_4}})\boldsymbol{{\rm f}}_s, \qquad \boldsymbol{{\rm f}}’_s=[\boldsymbol{{\rm f}}_s, \boldsymbol{{\rm z}}_s], \\
&\boldsymbol{{\rm z}}_t = {\rm softmax}(\frac{\boldsymbol{{\rm f}}_t\boldsymbol{{\rm f}}_s^T}{\sqrt{C_4}})\boldsymbol{{\rm f}}_t, \qquad \boldsymbol{{\rm f}}’_t=[\boldsymbol{{\rm f}}_t, \boldsymbol{{\rm z}}_t].
\end{align}
\vskip -0.05in

The output of this block is the concatenation of input features and attended features. Then, we upsample $\boldsymbol{{\rm f}}'_s, \boldsymbol{{\rm f}}'_t$ to get dense features $\boldsymbol{{\rm f}}^d_s, \boldsymbol{{\rm f}}^d_t \in \mathbb{R}^{N_1 \times C_5}$ using a PointNet++ decoder. 
Following the principle in the above, we convert these dense point features into the keypoint saliency volumes by projection and a 3D UNet $U_k$.
Suppose the full detection module is denoted as $\mit \Psi$, the outputs are named as $\mit\Psi(\boldsymbol{{\rm o}}_s, \boldsymbol{{\rm o}}_t)_s, \mit\Psi(\boldsymbol{{\rm o}}_s, \boldsymbol{{\rm o}}_t)_t \in \mathbb{R}^{m \times C_h \times C_w \times C_d}$.
Then, we can marginalize (see section 3.1 of \cite{kulkarni2019unsupervised}) the saliency volumes along three orthogonal axes to extract $m$ 3D keypoints $\boldsymbol{{\rm k}}_s, \boldsymbol{{\rm k}}_t \in \mathbb{R}^{m \times 3}$, as visualized in the blue panel in Fig. \ref{fig:main}. Here, the $i$-th keypoint in $\boldsymbol{{\rm k}}_s$ and $\boldsymbol{{\rm k}}_t$ correspond to each other ($i \in [1,m]$).


\textbf{Feature Transportation.} Similar to 2D Transporter, the next step involves feature transportation for reconstructing $\boldsymbol{{\rm o}}_t$ from $\boldsymbol{{\rm o}}_s$.
We transport the features in $\mit \Phi(\boldsymbol{{\rm o}}_t)$ around $\boldsymbol{{\rm k}}_t$ into $\mit \Phi(\boldsymbol{{\rm o}}_s)$ and suppress features in $\mit \Phi(\boldsymbol{{\rm o}}_s)$ around $\boldsymbol{{\rm k}}_t$ and $\boldsymbol{{\rm k}}_s$. As shown in the green panel in Fig. \ref{fig:main}, we first erase features at both sets of keypoints in $\mit \Phi(\boldsymbol{{\rm o}}_{s})$ to get $\mit \Phi^-(\boldsymbol{{\rm o}}_{s})$, then extract the feature surrounding keypoints $\boldsymbol{{\rm k}}_t$ from $\boldsymbol{\rm o}_{t}$, and finally combine both to generate $\mit \Phi^+(\boldsymbol{{\rm o}}_{s})$, which is formulated by:
\vspace{-1mm}
\begin{align}
\label{equ:transport}
\mit \Phi^+(\boldsymbol{{\rm o}}_{s}) = &(1-H_{\mit \Psi(\boldsymbol{{\rm o}}_s, \boldsymbol{{\rm o}}_t)_s}) \cdot (1-H_{\mit \Psi(\boldsymbol{{\rm o}}_s, \boldsymbol{{\rm o}}_t)_t}) \cdot \mit \Phi(\boldsymbol{{\rm o}}_{s}) \nonumber
\\ &+ H_{\mit \Psi(\boldsymbol{{\rm o}}_s, \boldsymbol{{\rm o}}_t)_t} \cdot \mit\Phi(\boldsymbol{{\rm o}}_{t})\text{,}
\end{align}
where $H_{\mit\Psi}$ represents a 3D heatmap made out of fixed-variance $\sigma$ isotropic Gaussians centered at each of the $m$ keypoint coordinates indicated by $\mit\Psi$, and $(1-H_{\mit \Psi(\boldsymbol{{\rm o}}_s, \boldsymbol{{\rm o}}_t)_s}) \cdot (1-H_{\mit \Psi(\boldsymbol{{\rm o}}_s, \boldsymbol{{\rm o}}_t)_t}) \cdot \mit\Phi(\boldsymbol{{\rm o}}_{s})$ is denoted as $\mit\Phi^-(\boldsymbol{{\rm o}}_{s})$ in Fig. \ref{fig:main}.

It's worth noting that the exact values of feature dimensions in our experiments are detailed in the supplementary.

\subsubsection{Geometry Implicit Decoder}
Since the geometry except for moving parts remains the same between the source and target frame, building transported features using detected corresponding keypoints of moving parts can enable the re-synthesis of the target visual inputs.
As the 2D Transporter does not change the data structure after transportation, it is easily achievable to reconstruct the input image by a CNN-based decoder. This is not viable for irregular 3D data,
 so our 3D Transporter utilizes implicit neural representations to reconstruct the underlying shape of the target instead of the raw point clouds. This is motivated by recent studies demonstrating the effectiveness of deep implicit functions for 3D reconstruction. 
By mapping the irregular point clouds into volumetric features, we find that using implicit shape decoding is more effective compared to sparse reconstruction (see Tab.~\ref{tab:ablation}).

Given a point $\boldsymbol{{\rm q}} \in \mathbb{R}^3$ from a query set $Q$, our method encodes it into a ${C_e}$-dimensional vector $\boldsymbol{{\rm q}}_e$ using a multi-layer perceptron. Then, the local feature $\mit \Phi_{\boldsymbol{{\rm q}}}^{+}(\boldsymbol{{\rm o}}_s)$ is queried from the transported feature volume $\mit \Phi^+(\boldsymbol{{\rm o}}_{s})$ via trilinear interpolation. Our implicit decoder ${\mit \Omega}$ maps the concatenation of feature $\boldsymbol{{\rm q}}_e$ and $\mit \Phi_{\boldsymbol{{\rm q}}}^{+}(\boldsymbol{{\rm o}}_s)$ to a target surface occupancy probability ${\rm Prob}(\boldsymbol{{\rm q}}|\boldsymbol{{\rm o}}_t) \in [0, 1]$, as formulated by: 
\vspace{-1mm}
\begin{align}
\label{equ:occp}
{\mit \Omega }(\boldsymbol{{\rm q}}_e, \mit \Phi_{\boldsymbol{{\rm q}}}^{+}(\boldsymbol{{\rm o}}_s)) \rightarrow {\rm Prob}(\boldsymbol{{\rm q}}|\boldsymbol{{\rm o}}_t)\text{.}
\end{align}

\subsubsection{Loss Function}
All modules could be optimized by a surface reconstruction loss. As we claim that we have no access to any information other than the given videos, we solely use the input point clouds for training the implicit decoder. Specifically, we define occupied points are those lying on the input surface, while all other points are considered unoccupied, including those inside and outside the surface. 

The binary cross-entropy loss between the predicted target surface occupancy ${\rm Prob}(\boldsymbol{{\rm q}}|\boldsymbol{{\rm o}}_t)$ and the ground-truth labels of the target frame ${\rm Prob}^{\rm gt}(\boldsymbol{{\rm q}}|\boldsymbol{{\rm o}}_t)$ is used. 
If $\boldsymbol{{\rm q}}$ is from the input target point clouds, the ${\rm Prob}(\boldsymbol{{\rm q}}|\boldsymbol{{\rm o}}_t)$ would be 1, otherwise be 0. 
We randomly sample queries $Q$ from the volume of size $C_h \times C_w \times C_d$ and the target point clouds, then average the results over all queries:
\begin{align}
\label{equ:occp_loss}
\mathcal{L}_{\rm occ\_t}=\frac{1}{|Q|}\sum_{\boldsymbol{{\rm q}}\in Q}l_{\rm BCE} \big({\rm Prob}(\boldsymbol{{\rm q}}|\boldsymbol{{\rm o}}_t), {\rm Prob}^{\rm gt}(\boldsymbol{{\rm q}}|\boldsymbol{{\rm o}}_t)\big),
\end{align} 
where $|Q|$ is the number of queries $Q$.

We also incorporate an additional loss term, $\mathcal{L}_{\rm occ\_s}$, to aid the source frame reconstruction process by leveraging its own feature grids, $\mit \Phi(\boldsymbol{{\rm o}}_{s})$. This loss term, formulated in the supplementary, leads to improved perception results.


\subsection{Manipulation using Consistent Keypoints}\label{sec: manipulation}
The use of keypoints as a mid-level representation of objects is an appropriate way for contact-rich robotic manipulation tasks that occur within a 3D space, such as tool manipulation \cite{qin2020keto}, object grasp \cite{manuelli2022kpam}, cloth folding \cite{ma2022learning} and generic visuomotor policy learning \cite{florence2019self}. However, previous works either focus solely on 2D keypoint representation or struggle to detect temporally consistent 3D keypoints when faced with shape variations and changes in object topology. Owing to the long-term consistency of 3D Transporter keypoints, our method is well-suited for handling 3D manipulation tasks. 
To demonstrate that, we choose articulated object manipulation as the benchmark.

The task is formulated in UMPNET \cite{xu2022umpnet}: Given a goal state $\boldsymbol{\rm o}_g$, a robot with an end-effector aims to generate a set of actions by which the articulated object can be moved from the current state $\boldsymbol{\rm o}_c$ to $\boldsymbol{\rm o}_g$, as shown in Fig.~\ref{fig:action_rule}. In this study, each state is in the form of point clouds instead of RGB-D images as used in \cite{xu2022umpnet}. Here, we use a suction-based gripper that can grasp any point on the object surface as \cite{EisnerZhang2022FLOW,xu2022umpnet} used.

Before manipulation, we leverage the geometry prior about articulation to design two additional losses during keypoint learning for improving the performance of 3D Transporter keypoint estimates:

\textbf{Keypoint Correspondence Loss}
Since the predicted keypoints are expected to scatter on the mobile part, we can use them to generate the pose hypothesis of the rigid part motion between the source and target, which is given by:
\vspace{-5mm}
\begin{align}
\mathbf{\hat{R}}, \mathbf{\hat{t}}=\underset{\mathbf{R}, \mathbf{t}}{\min } \sum_{i=1}^m\left\|\boldsymbol{{\rm k}}_g^i-\left(\mathbf{R} \cdot \boldsymbol{{\rm k}}_{c}^i+\mathbf{t}\right)\right\|^2\text{.}
\label{equ:svd}
\end{align}
This can be computed in closed form using SVD \cite{besl1992method}.
We enforce all correspondent keypoints to meet this rigid transformation to make keypoints geometrically aligned:
\begin{align}
\mathcal{L}_{\rm corr}= \sum_{i=1}^m \left\|\boldsymbol{{\rm k}}_g^i-\left(\mathbf{\hat{R}} \cdot \boldsymbol{{\rm k}}_{c}^i+\mathbf{\hat{t}}\right)\right\|^2\text{.}
\end{align}

\textbf{Joint Consistent Loss}
During an articulated motion, we note that the axis orientation of a certain joint under different states should remain the same or parallel.
Given the predicted pose by Eq. (\ref{equ:svd}), we can calculate its axis orientation $\boldsymbol{\upmu}$ and angle $\theta$, through Rodrigues' rotation formula. We penalize the difference in axis orientation of a certain joint at different time steps:
\begin{align}
\mathcal{L}_{\rm axis}=\min (1-\boldsymbol{\upmu}_{12}{\boldsymbol{\upmu}_{23}}^T, 1+\boldsymbol{\upmu}_{12}{\boldsymbol{\upmu}_{23}}^T)\text{,}
\end{align}
where $\boldsymbol{\upmu}_{ab}$ is the predicted axis orientation between observations at time-step $a$ and $b$.

The overall training loss is:
\begin{align}
\mathcal{L}=\mathcal{L}_{\rm occ\_t}+\mathcal{L}_{\rm occ\_s}+\lambda_1\mathcal{L}_{\rm corr}+\lambda_2\mathcal{L}_{\rm axis} \text{,}
\end{align}
where $\lambda_1$ and $\lambda_2$ are the loss weights (exact values are detailed in the supplementary).

After training, we develop an object-agnostic manipulation policy based on 3D Implicit Transporter keypoints, avoiding the notoriously inefficient exploration used in \cite{xu2022umpnet}. 
We first define each action $\boldsymbol{\rm A}_{c,g}$, that moves the object from its \textbf{c}urrent state to the \textbf{g}oal state,
as a 6-Dof pose which indicates the suction position $\boldsymbol{\rm A}^{pos}_{c,g} \in \mathbb{R}^{3}$ and moving direction $\boldsymbol{\rm A}^{dir}_{c,g} \in \mathbb{R}^{3}$. Then, our policy consists of two parts:

\begin{figure}[t]
    \centering
    \includegraphics[width=0.45\textwidth]{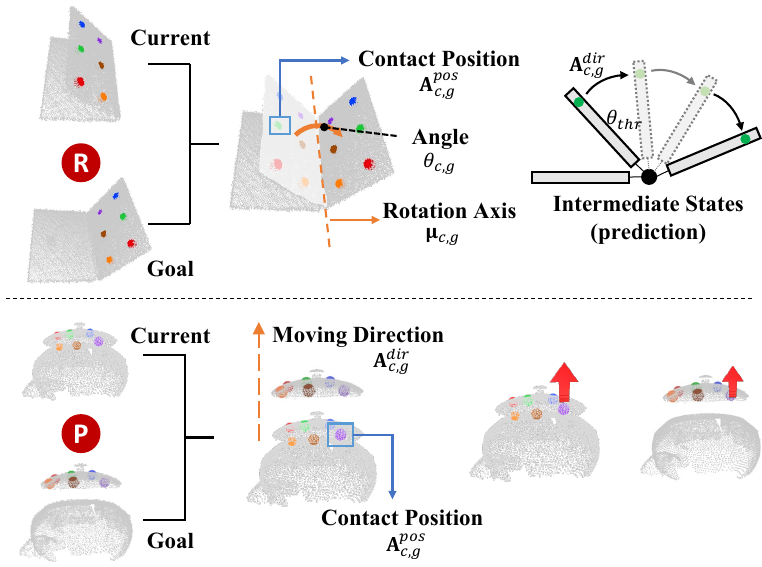}
    \caption{Illustrations of our manipulation policy based on correspondent keypoints. R means revolute and P means prismatic. The length of the red arrow is proportional to the action distance.}
    \label{fig:action_rule}
    \vspace{-4mm}
\end{figure}

\textbf{Position and Direction Inference}
The first step is to obtain predicted keypoints $\boldsymbol{{\rm k}}_c, \boldsymbol{{\rm k}}_g$, axis ${\boldsymbol{\upmu}}_{c,g}$ and angle $\theta_{c,g}$.
Then, we compute the sparse articulation flow $\boldsymbol{\rm F}_{c,g}^i=\boldsymbol{\rm k}_g^i-\boldsymbol{\rm k}_c^i$ from correspondent keypoints. 
To efficiently actuate the moving part, we select the keypoint location $\boldsymbol{\rm k}_c^{s}$ with the highest magnitude flow as the \textbf{s}uction point according to \emph{the principle of leverage}, denoted as $\boldsymbol{\rm A}^{pos}_{c,g}=\boldsymbol{\rm k}_c^{s}$. 

For a revolute joint (Fig.~\ref{fig:action_rule} top panel), $\boldsymbol{\rm k}_c^{s}$ is restricted to move on a 2D circle with a radius $\boldsymbol{\rm r}^s$ perpendicular to the axis of rotation, where $\boldsymbol{\rm r}^s$ is the shortest vector from $\boldsymbol{\rm k}_c^{s}$ to the joint link. Therefore, the ideal action direction $\boldsymbol{\rm A}^{dir_s}_{c,g}$ for $\boldsymbol{\rm k}_c^{s}$ is tangent to the circle formed by $\boldsymbol{\rm r}^s$. 
If the range of motion between the current and goal state is small, w.r.t. $\theta_{c,g} \leq \theta_{thr}$, 
where $\theta_{thr}$ is a threshold, $\boldsymbol{\rm F}_{c,g}^{s}$ is approximately parallel to $\boldsymbol{\rm A}^{dir_s}_{c,g}$, which can be set as the action direction. However, when the rotation is increased, the difference between them gets more significant. This issue can be alleviated by interpolating some intermediate states according to the predicted axis. Specifically, we use the axis $\boldsymbol{\upmu}_{c,g}$ and $\theta_{thr}$ to rotate $\boldsymbol{\rm k}_{c}^{s'}$ (the result of moving center of $\boldsymbol{\rm k}_c^{s}$ to origin coordinates) to generate $\boldsymbol{\rm k}_{c_1}^{s'}$. Then $\boldsymbol{\rm F}_{c,c_1}^{s'}$ can be computed as the action direction at $\boldsymbol{\rm k}_{c}^{s}$.

For a prismatic joint (Fig.~\ref{fig:action_rule} bottom panel), $\boldsymbol{\rm A}^{dir_s}_{c,g}$ is parallel to the ground-truth articulation flow of each moving point. Therefore, we can directly use $\boldsymbol{\rm F}_{c,g}^{s}$ as the articulate direction. Since the motion of a prismatic joint always satisfies the condition $\theta_{c,g}\leq\theta_{thr}$, and the rule for direction estimates is the same as the revolute joint, \textbf{there is no need to classify the joint type}.

\textbf{Closed-loop Manipulation}
In contrast to using a single-step action to reach the target, we generate a sequence of actions over multiple steps to gradually change the articulation state. To achieve this, we adopt a closed-loop control system that relies on feedback to adjust the current action.  
Specifically, we predict the next action based on the object’s current and goal state. 
Unlike the previous method \cite{xu2022umpnet} that used a constant moving distance,
we leverage the magnitude of articulation flow $\boldsymbol{\rm F}_{c,g}^s$ to adjust the moving distance dynamically. 
Inspired by the idea of a PID controller \cite{ang2005pid}, we set $||\boldsymbol{\rm A}^{dir_s}_{c,g}||=\lambda \cdot ||\boldsymbol{\rm F}_{c,g}^{s}||$ at the current state, where $\lambda$ is a proportional coefficient. 


\section{Experiments}
We demonstrate the effectiveness of our approach in both perception and manipulation tasks. We begin with an evaluation of 3D correspondent keypoint detection methods on both synthetic and real-world datasets. Additionally, we conduct an ablation study to investigate the impact of each design choice in our approach. Next, we demonstrate our method's ability to perform goal-conditioned manipulation in simulation and further verify our approach's practicality by showcasing its performance on a real-world platform. More details regarding our implementation and hyper-parameters can be found in the supplementary.

\begin{table*}[t]
\caption{The perception performance of ours and other baselines.}
\label{tab:perception}
\centering
{\scriptsize 
    \setlength\tabcolsep{2pt}
    \begin{tabular}{l|ccccccccccc|cccccccccc}
    \hline
      & \multicolumn{11}{c|}{\textbf{Novel Instances in Train Categories}} &  \multicolumn{10}{c}{\textbf{Test Categories}}\\
     & 
    \includegraphics[width = 0.032\linewidth]{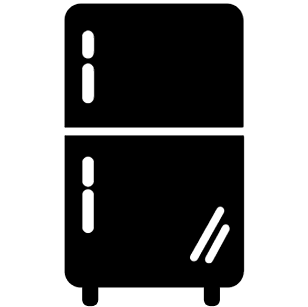} &
    \includegraphics[width = 0.032\linewidth]{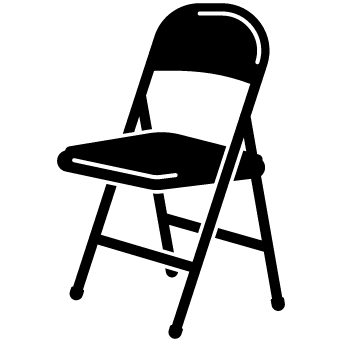} & 
    \includegraphics[width = 0.032\linewidth]{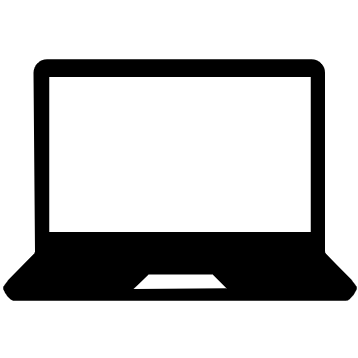} &
    \includegraphics[width = 0.032\linewidth]{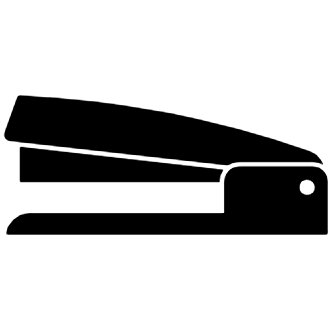} &
    \includegraphics[width = 0.032\linewidth]{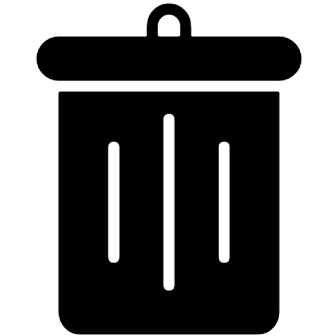} &
    \includegraphics[width = 0.032\linewidth]{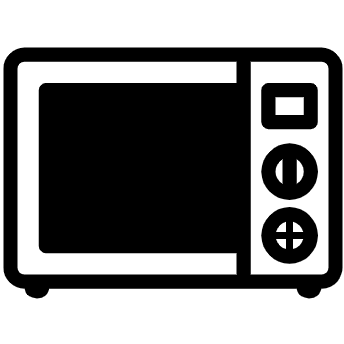} &
    \includegraphics[width = 0.032\linewidth]{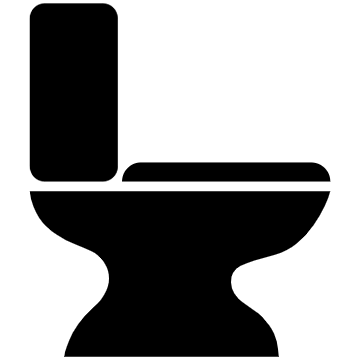} &
    \includegraphics[width = 0.032\linewidth]{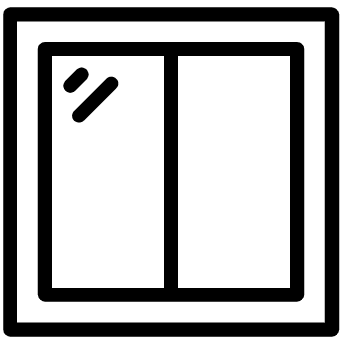} &
    \includegraphics[width = 0.032\linewidth]{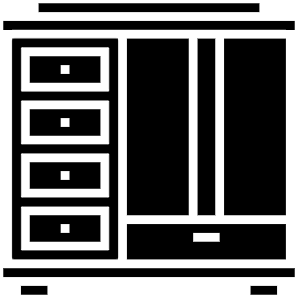} &
    \includegraphics[width = 0.032\linewidth]{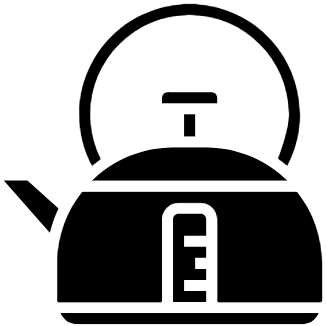} &
    \rotatebox{60}{\textbf{Avg.}} &
    \includegraphics[width = 0.032\linewidth]
    {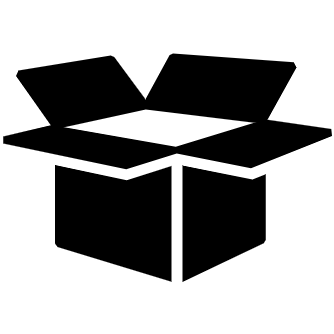} &
    \includegraphics[width = 0.032\linewidth]{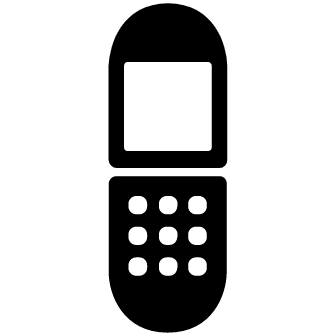} &
    \includegraphics[width = 0.032\linewidth]{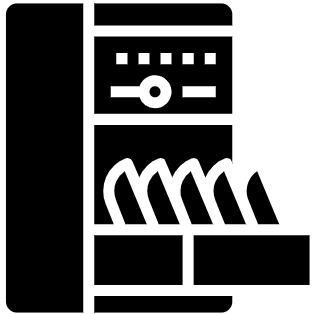} &
    \includegraphics[width = 0.032\linewidth]{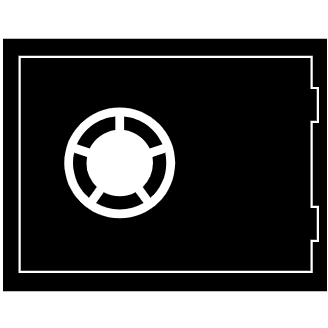} &
    \includegraphics[width = 0.032\linewidth]{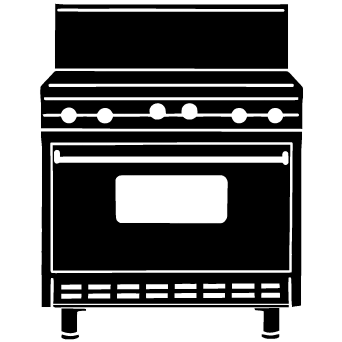} &
    \includegraphics[width = 0.032\linewidth]{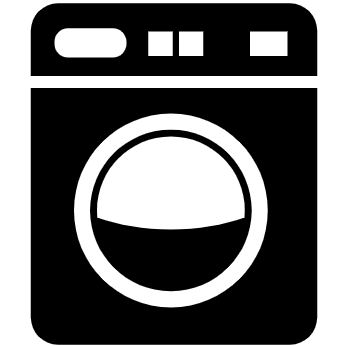} &
    \includegraphics[width = 0.032\linewidth]{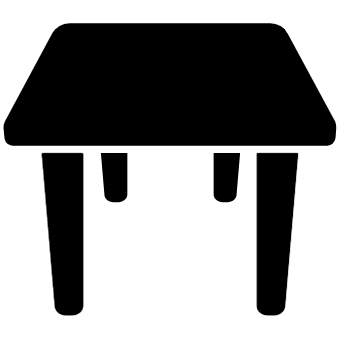} &
    \includegraphics[width = 0.032\linewidth]{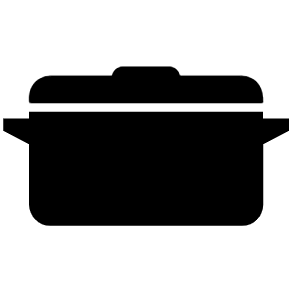} &
    \includegraphics[width = 0.032\linewidth]{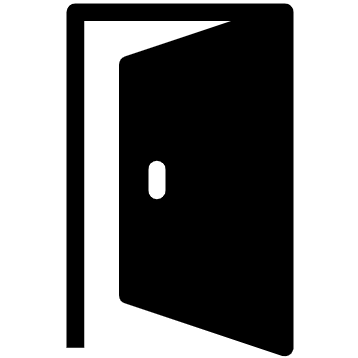} &
    {\rotatebox{60}{\textbf{Avg.}}}
    \\ \hline
    
    \multicolumn{22}{c}{\cellcolor[HTML]{DEEBF7}\textbf{Average correspondent keypoint distance (ACKD) $\downarrow$}} \\ \hline
    Random & 0.24 & 0.26 & 0.34 & 0.31 & 0.27 & 0.24 & 0.21 & 0.28 & 0.24 & 0.13 & 0.25 & 0.32 & 0.30 & 0.32 & 0.25 & 0.25 & 0.18 & \textbf{0.21} & 0.24 & 0.31 & 0.26\\
    
    ISS\cite{zhong2009intrinsic} & 0.26 & 0.24 & 0.31 & 0.28 & 0.26 & 0.24 & 0.20 & 0.25 & 0.25 & 0.14 & 0.24 & 0.31 & 0.29 & 0.33 & 0.23 & 0.26 & 0.19 & 0.22 & 0.20 & 0.30 & 0.26\\
    
    USIP\cite{li2019usip} & 0.23 & 0.28 & 0.34 & 0.36 & 0.26 & 0.23 & 0.25 & 0.24 & \textbf{0.23} & 0.13 & 0.26 & 0.33 & 0.34 & 0.38 & 0.19 & 0.29 & 0.21 & 0.23 & 0.21 & 0.26 & 0.27\\
    
    SNAKE\cite{zhong2022snake} & 0.25 & 0.26 & 0.31 & 0.29 & 0.25 & 0.28 & 0.21 & 0.27 & 0.25 & 0.16 & 0.25 & 0.29 & 0.31 & 0.31 & 0.24 & 0.24 & 0.18 & 0.22 & 0.26 & 0.29 & 0.26\\
    
    D3feat\cite{bai2020d3feat} & \textbf{0.20} & 0.21 & 0.22 & 0.21 & 0.25 & 0.23 & \textbf{0.11} & 0.24 & 0.25 & 0.14 & 0.21 & 0.19 & 0.22 & 0.24 & 0.17 & 0.20 & 0.13 & \textbf{0.21} & 0.16 & 0.19 & 0.19\\
    
    Ours & 0.21 & \textbf{0.07} & \textbf{0.07} & \textbf{0.11} & \textbf{0.15} & \textbf{0.11} & \textbf{0.11} & \textbf{0.14} & \textbf{0.23} & \textbf{0.06} & \textbf{0.13} & \textbf{0.13} & \textbf{0.06} & \textbf{0.12} & \textbf{0.15} & \textbf{0.18} & \textbf{0.12} & 0.22 & \textbf{0.07} & \textbf{0.13} & \textbf{0.13}\\ 
    \hline
    
    
    
    
    
    
    
    \multicolumn{22}{c}{\cellcolor[HTML]{DEEBF7}{\textbf{Average distance on pose estimation (ADD) $\downarrow$}}}\\ \hline
    Random & 0.26 & 0.26 & 0.31 & 0.34 & 0.27 & 0.27 & 0.21 & 0.28 & 0.26 & 0.18 & 0.26 & 0.31 & 0.29 & 0.34 & 0.28 & 0.26 & 0.18 & 0.22 & 0.32 & 0.33 & 0.28\\
    
    ISS\cite{zhong2009intrinsic} & 0.27 & 0.24 & 0.32 & 0.30 & 0.26 & 0.28 & 0.22 & 0.23 & 0.26 & 0.18 & 0.26 & 0.29 & 0.30 & 0.35 & 0.24 & 0.26 & 0.19 & 0.22 & 0.30 & 0.33 & 0.27\\
    
    USIP\cite{li2019usip} & 0.28 & 0.25 & 0.33 & 0.38 & 0.26 & 0.27 & 0.23 & 0.27 & 0.27 & 0.17 & 0.27 & 0.29 & 0.33 & 0.38 & 0.25 & 0.28 & 0.24 & 0.24 & 0.32 & 0.30 & 0.29\\
    
    SNAKE\cite{zhong2022snake} & 0.26 & 0.25 & 0.33 & 0.33 & 0.28 & 0.29 & 0.24 & 0.24 & 0.26 & 0.19 & 0.27 & 0.29 & 0.32 & 0.37 & 0.24 & 0.26 & 0.21 & \textbf{0.21} & 0.37 & 0.29 & 0.29\\
    
    D3feat\cite{bai2020d3feat} & 0.21 & 0.22 & 0.26 & 0.24 & 0.27 & 0.27 & 0.16 & 0.24 & 0.28 & 0.19 & 0.23 & 0.21 & 0.26 & 0.34 & 0.19 & 0.27 &0.16 & \textbf{0.21} & 0.21 & 0.18 & 0.23\\
    
    Ours & \textbf{0.18} & \textbf{0.08} & \textbf{0.07} & \textbf{0.10} & \textbf{0.16} & \textbf{0.11} & \textbf{0.11} & \textbf{0.13} & \textbf{0.21} & \textbf{0.07} & \textbf{0.12} & \textbf{0.14} & \textbf{0.06} & \textbf{0.11} & \textbf{0.15} & \textbf{0.15} & \textbf{0.12} & 0.24 & \textbf{0.07} & \textbf{0.12} & \textbf{0.13}
    \\ \hline    
    
    \end{tabular}
    \begin{tablenotes}
        \item[1] Classes: fridge, folding chair, laptop, stapler, trashcan, microwave, toilet, window, cabinet, kettle, box, phone, dishwasher, safe, oven, washing machine, table, kitchen pot, door.
    \end{tablenotes}
}
\end{table*}

\subsection{Datasets}
\textbf{PartNet-Mobility} \cite{xiang2020sapien}: We adopt a similar approach to Xu \textit{et al.} \cite{xu2022umpnet} for selecting synthetic object models from PartNet-Mobility to generate our data, except for two categories with an insufficient number of instances. Thus, we train on 10 categories and test on 9 categories, with specific object classes listed in the footnote of Tab.~\ref{tab:perception}. For the perception task, we load one instance at a time into the Pybullet simulator with random pose and joint configurations, and gradually change the articulation states of a randomly selected joint to mimic human-object interaction. For the manipulation task, we use the same joint configurations of testing objects at the initial and goal states as \cite{xu2022umpnet} set. At each state in both tasks, we integrate three-view rendered depth images into point clouds.
Examples of generated data are visualized in the supplementary. 


\textbf{ITOP} \cite{haque2016towards}: The ITOP dataset, which consists of depth map sequences capturing diverse real human actions, is used in our perception task. Specifically, we select 9k training frames and 1k testing frames from the dataset.

\textbf{Rodent3D} \cite{patel2023animal}: The Rodent3D dataset contains 240 minutes of multimodal (RGB, depth, and thermal) video recordings depicting rodents exploring an arena in a laboratory. The dataset is leveraged to develop a model aimed at accurately tracking the 3D pose of animals.

\subsection{Baselines}
For the perception task, we compare our method with several 3D keypoint detection methods, which contains 
random guess, hand-crafted detectors:
ISS \cite{zhong2009intrinsic}, and deep learning-based unsupervised detectors: USIP \cite{li2019usip} and SNAKE \cite{zhong2022snake}. To find correspondence between keypoints of two observations, we need both keypoint detectors and descriptors. As such, we use an off-the-shelf and generic descriptor FPFH \cite{rusu2009fast} with the abovementioned keypoint detectors. Moreover, the matching is based on the nearest neighbor search. We also choose a joint learning method for 3D keypoint detection and description: D3feat \cite{bai2020d3feat}. All the baselines are pre-trained on our dataset.

For the manipulation task, UMPNet \cite{xu2022umpnet} is selected as a strong baseline, which proposed to use a universal strategy to handle various objects for goal-conditioned manipulation. We also compare with a single-step action model proposed by Agrawal \textit{et al.} \cite{agrawal2016learning}.

\begin{figure}[t]
    \centering
    \includegraphics[width=0.45\textwidth]{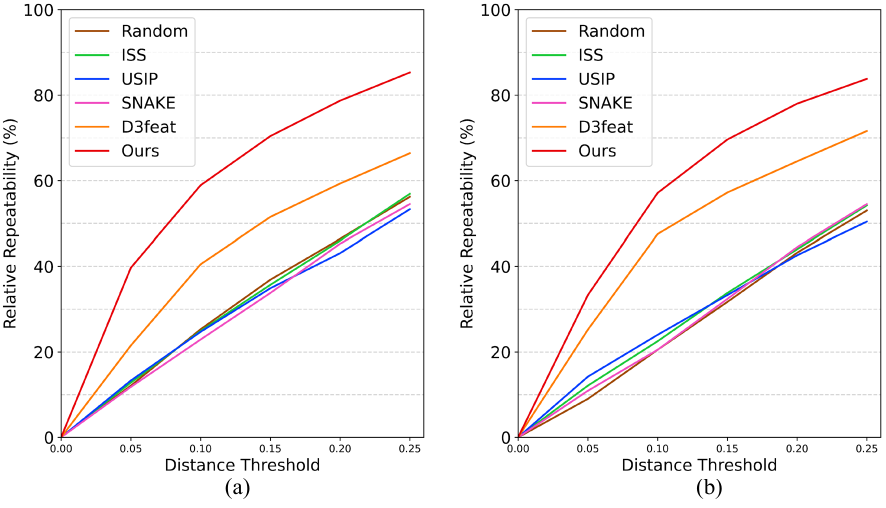}
    \vspace{-2mm}
    \caption{Relative repeatability (\%) of novel instances in train categories (a) and test categories (b). Our method significantly surpasses other baselines. Note that our counterparts need a segmentation mask to filter the keypoints on the mobile part.}
    \label{fig:repeatability}
    \vspace{-5.0mm}
\end{figure}

\subsection{Correspondent Keypoint Detection}

\begin{figure*}[ht]
  \centering
  \includegraphics[width=0.9\linewidth]{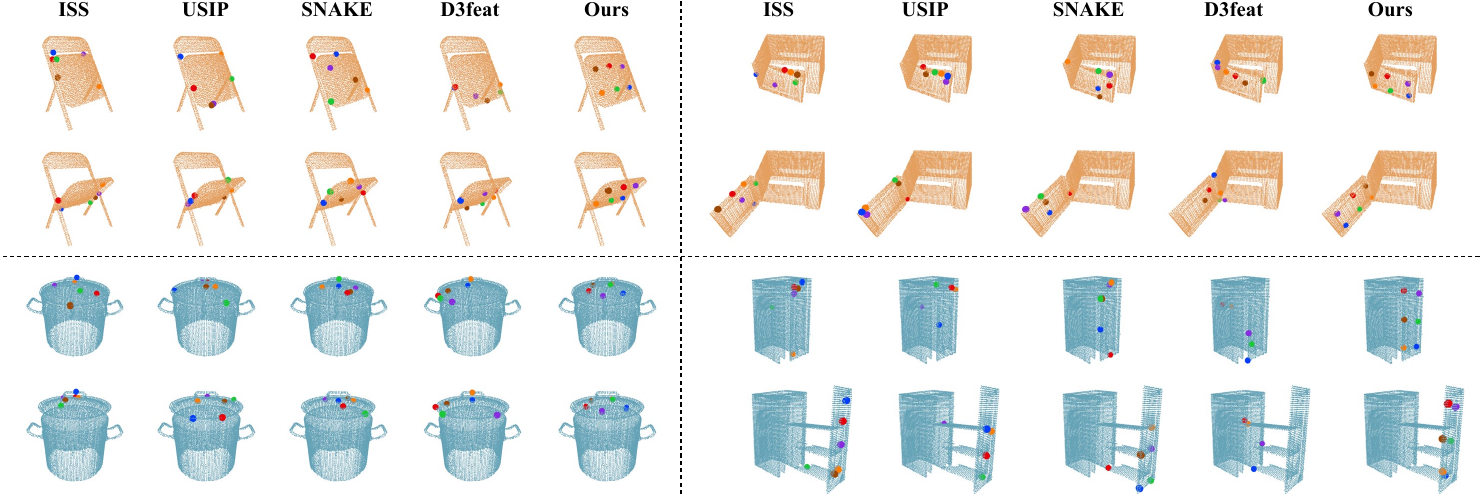}
  \caption{Temporal consistency of keypoints on articulated objects in different articulation states. The object in yellow has a revolute joint, and the blue one has a prismatic joint. The keypoint in the same color is correspondent.}
  \label{fig:kp_vis}
  \vskip -0.1 in
\end{figure*}

\begin{table}[!ht]
\caption{Results on the ITOP dataset under two settings: w/ and w/o a mask to filter keypoints of the human body at the test.}
\centering
\resizebox{0.45\textwidth}{!}{
\begin{tabular}{c|cccccc}
    \toprule
       & Random & ISS & USIP & SNAKE & D3feat & Ours \\
    \midrule
    w/ mask & 0.53 & 0.29 & 0.23  & 0.27  & 0.22  & \textbf{0.13} \\
    w/o mask & 0.74 & 0.89 & 0.87  & 0.69  & 0.78  & \textbf{0.14} \\
    \bottomrule
\end{tabular}}%
\label{tab:human}
\vskip -0.1in
\end{table}

\begin{figure}[!ht]
 \centering
\includegraphics[height=0.6\linewidth]{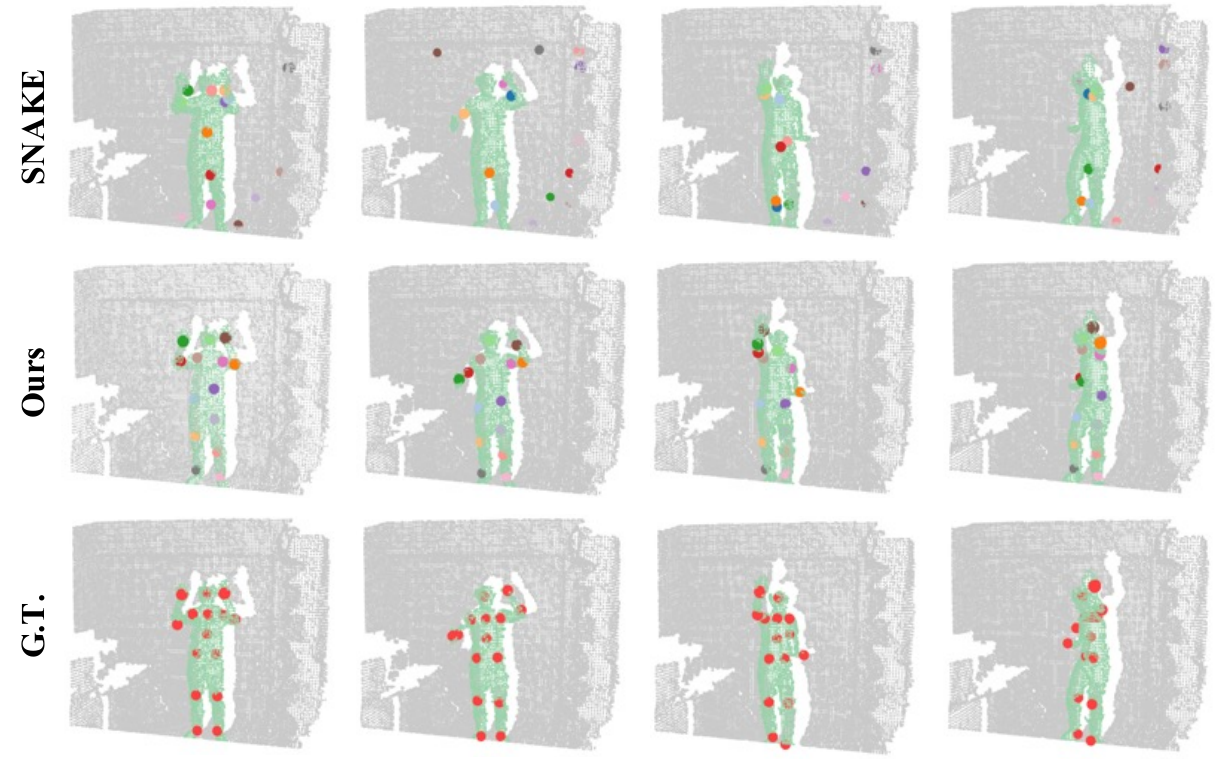}
\caption{Keypoint consistency on an articulated human.}
\label{fig:human}
\vskip -0.2in
\end{figure}

%

\vspace{-2.0mm}
\begin{figure}[!ht]
  \centering
\includegraphics[width=0.43\textwidth]{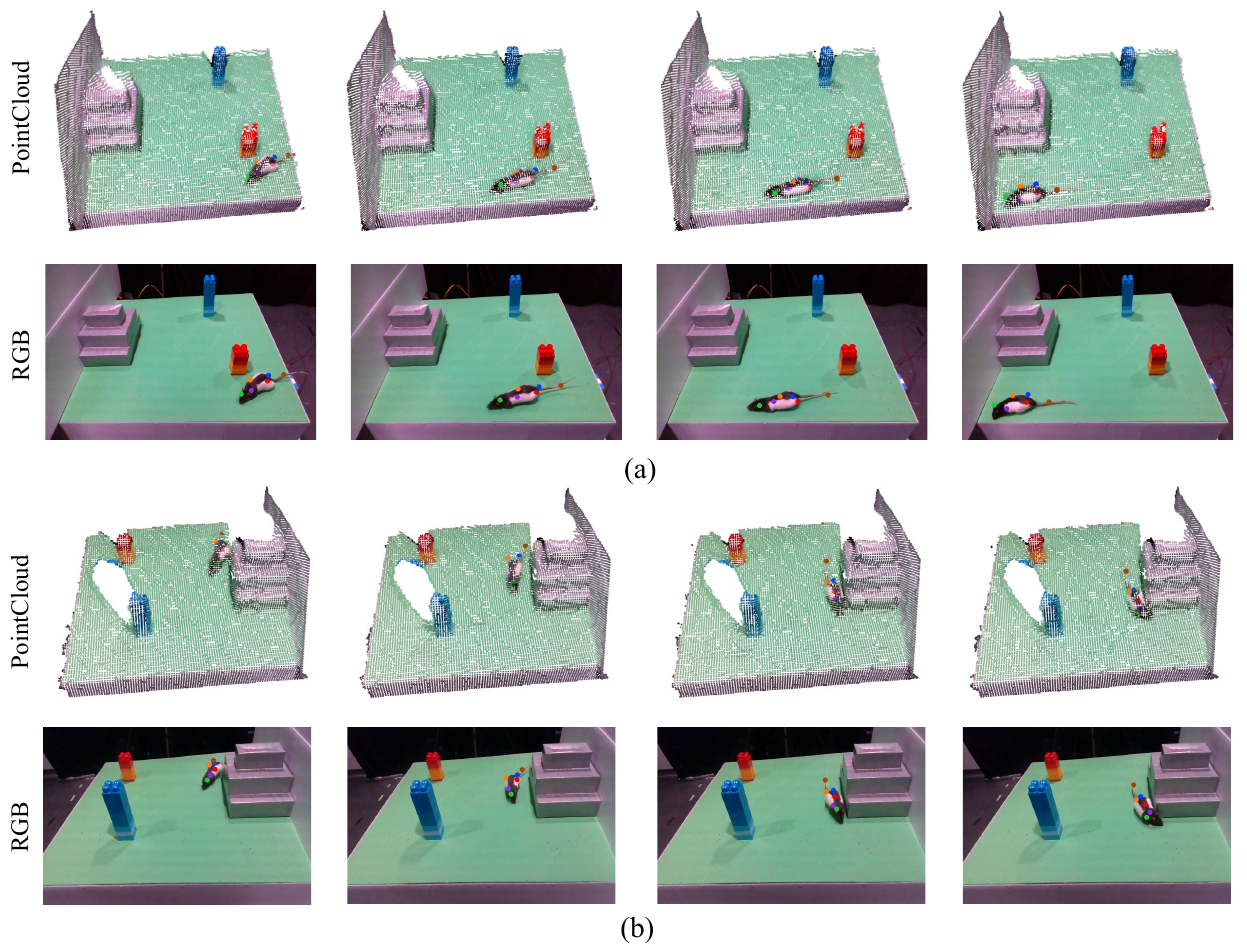}
\vspace{-1.0mm}
  \caption{Keypoint predictions of ours on the Rodent3D dataset.}
  \label{fig:rodent3d}
  \vspace{-6.0mm}
\end{figure}

\begin{figure*}[t]
  \centering
  \includegraphics[width=0.9\linewidth]{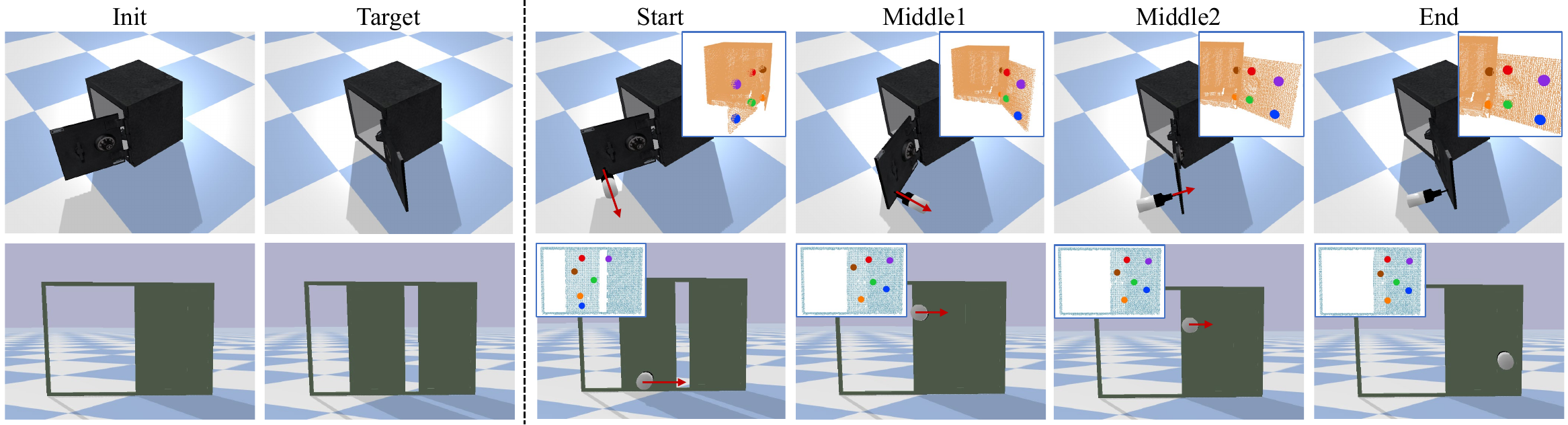}
  \caption{Visualizations of the goal-conditioned task, and the perception and manipulation results of our method in each step.}
  \label{fig:goal conditioned manipulation}
  \vspace{-4.0mm}
\end{figure*}

We compare the keypoint consistency of two different articulation states of the same instance. We mainly focus on keypoints of the moving part to demonstrate the temporal consistency. Note that we need to use the ground-truth segmentation mask to select keypoints to meet the above requirement for all baselines. For a fair comparison, we set a fixed keypoint number $m$ of 6 for each method and ours. \\
\textbf{Metrics}
We exploit the following metrics for evaluation:
1) \textit{Average correspondent keypoint distance} (ACKD): CKD is the Euclidean distance of correspondent keypoints in the same coordinate system. ACKD is the average CKD of all keypoints. 
2) \textit{Relative repeatability} (RR): A keypoint is repeatable if its CKD is below a distance threshold. RR is the percentage of repeatable points in a total number of detected keypoints. 
3) \textit{Average distance on pose estimation} (ADD): 
According to Eq. (\ref{equ:svd}), the part motion of an articulated object can be solved. Following PoseCNN \cite{xiang2017posecnn}, we adopt ADD to compare the distance between predicted pose [$\mathbf{\hat{R}}$, $\mathbf{\hat{t}}$] and ground-truth pose [$\mathbf{R}$, $\mathbf{t}$]. 
\\
\textbf{Evaluation Results}
The quantitative results on the \textit{PartNet-Mobility} dataset are provided in Tab.~\ref{tab:perception} and Fig.~\ref{fig:repeatability}. 
Our method demonstrates superior performance on both novel instances within the training set and test categories.
Baselines perform poorly on temporal keypoint detection. The reason may be twofold. 1) Their core principle is to find repeatable keypoints under view variations, which makes sense only for rigid objects. And they did not design a strategy to adapt to significant shape disturbances of the same object. 2) The bottleneck is the mismatch between keypoint detector and descriptor. The performance gap between the D3feat and other baselines is the apparent evidence to verify that. Owing to the design of the 3D feature transport and implicit shape reconstruction, our method can leverage object motion to discover keypoints scattered on the mobile parts without segmentation masks, which is different from all baselines. As shown in Fig.~\ref{fig:kp_vis}, our keypoints show good temporal alignment compared to other methods. 


We compute the Chamfer Distance for the \textit{ITOP} dataset using human-annotated semantic points, and the results are listed in Tab.~\ref{tab:human}. Note that all methods do not have access to masks of the human body points during training. Despite this, our proposed method achieves satisfactory performance in real-world scenarios compared to the baselines. The counterparts cannot distinguish keypoints on the moving body from those in the background without segmentation masks. The visualization in Fig.~\ref{fig:human} illustrates that our keypoints are close to the human labels.

Notably, the Rodent3D dataset does not furnish valid pose annotations, and therefore, we restrict ourselves to presenting qualitative outcomes of our 3D Implicit Transporter, as shown in Fig.~\ref{fig:rodent3d}. Our results demonstrate that the keypoint predictions generated by our method effectively capture the rodent's skeletal structure and display spatiotemporal coherence. Overall, our self-supervised method has the potential to significantly improve behavior analysis by addressing the difficulty of obtaining accurate 3D keypoint annotations for animals. More intuitive performance can be found in the supplementary video.

\textbf{Ablation Studies} 
1) \textit{Network and loss functions.} Tab.~\ref{tab:ablation} shows the average perception performance on test categories from the PartNet-Mobility dataset w.r.t. designs of our method. (Row 1-2) Using a point-based decoder like TopNet \cite{tchapmi2019topnet} to recover the target point cloud, rather than employing the implicit reconstruction results in a decrease in performance.
(Row 2-3) It shows that the keypoint performance improves when we utilize the cross-attention for feature fusion. (Row 3-5) The two loss functions are essential to make keypoints temporally consistent. 
2) \textit{Keypoint parameters.}
The keypoint number $m$ and Gaussian variance $\sigma$ are involved with the spatial range of transported features. 
The results for different settings of both parameters are shown in supplementary, which indicates that $m=6$ and $\sigma=0.15$ are the best choice in our settings. 3) \textit{Volume size}. The higher volumetric resolution of feature grids improves keypoint detection performance, as shown in the supplementary. However, this comes at the cost of increased memory usage. So we choose the voxel size of 64 to balance the memory cost and keypoint performance. 4) \textit{Query sampling}. Increasing the number of queries can improve the performance of keypoint detection as shown in Tab. \ref{tab:sampling_strategy}. But this increases the cost of memory and training time as well. Besides, it is crucial to sample negative queries randomly.


\vspace{-2mm}
\begin{table}[h]
\caption{Ablations for the designs of our method.}
\label{tab:ablation}
\begin{center}
\resizebox{0.9\columnwidth}{!}{
\begin{tabular}{cccc|ccc}
\hline
Imp. Rec. & Cro. Att. & $\mathcal{L}_{corr}$ & $\mathcal{L}_{axis}$ & RR $\uparrow$ & ACKD $\downarrow$ & ADD $\downarrow$ \\ \hline
&  &  &  & 0.170 & 0.286 & 0.317 \\
\rowcolor[HTML]{EFEFEF}
\checkmark &  &  &  & 0.457 & 0.199 & 0.186 \\
\checkmark & \checkmark &  &  & 0.530 & 0.153 & 0.143  \\
\rowcolor[HTML]{EFEFEF}
\checkmark & \checkmark & \checkmark & & 0.602 & 0.130 & 0.123\\
\checkmark & \checkmark & \checkmark & \checkmark & \textbf{0.611} & \textbf{0.127} & \textbf{0.109} \\ \hline
\end{tabular}
}
\end{center}
\vskip -0.3in
\end{table}

\vspace{-2mm}
\begin{table}[!ht]
\caption{Analysis of query sampling strategies. ’R’ and ’U’ means random and uniform sampling of negative queries.}
\centering
\resizebox{0.35\textwidth}{!}{
\begin{tabular}{c|ccc|c}
    \toprule
    Query \#  & 5k(R) & 2k(R) & 1k(R) & 2k(U) \\
    \midrule
    Repeatability $\uparrow$ & \textbf{0.663} & 0.611 & 0.597  & 0.058 \\
    \bottomrule
\end{tabular}}%
\label{tab:sampling_strategy}
\vskip -0.2in
\end{table}


\begin{table*}[t]
\caption{The manipulation performance of ours and other baselines.}
\label{tab:manipulation_results}
\centering
{\scriptsize 
    \setlength\tabcolsep{2pt}
    \begin{tabular}{l|ccccccccccc|cccccccccc}
    \hline
      & \multicolumn{11}{c|}{\textbf{Novel Instances in Train Categories}} &  \multicolumn{10}{c}{\textbf{Test Categories}}\\
     & 
    \includegraphics[width = 0.032\linewidth]{icon/noun_Fridge_1875643.png} &
    \includegraphics[width = 0.032\linewidth]{icon/noun_folding_chair_2151184.png} & 
    \includegraphics[width = 0.032\linewidth]{icon/noun_Laptop_2291662.png} &
    \includegraphics[width = 0.032\linewidth]{icon/noun_Stapler_2557851.png} &
    \includegraphics[width = 0.032\linewidth]{icon/noun_trashcan_2244926.png} &
    \includegraphics[width = 0.032\linewidth]{icon/noun_Microwave_1041630.png} &
    \includegraphics[width = 0.032\linewidth]{icon/noun_Toilet_3121.png} &
    \includegraphics[width = 0.032\linewidth]{icon/noun_window_3203560.png} &
    \includegraphics[width = 0.032\linewidth]{icon/noun_Cabinet_2881254.png} &
    \includegraphics[width = 0.032\linewidth]{icon/noun_Kettle_3002541.png} &
    \rotatebox{60}{\textbf{Avg.}} &
    \includegraphics[width = 0.032\linewidth]
    {icon/noun_Box_1650724.png} &
    \includegraphics[width = 0.032\linewidth]{icon/noun_flip_phone_143303.png} &
    \includegraphics[width = 0.032\linewidth]{icon/noun_dish_washer_3307528.png} &
    \includegraphics[width = 0.032\linewidth]{icon/noun_safe_1202915.png} &
    \includegraphics[width = 0.032\linewidth]{icon/noun_Oven_7255.png} &
    \includegraphics[width = 0.032\linewidth]{icon/noun_Laundry_1976992.png} &
    \includegraphics[width = 0.032\linewidth]{icon/noun_Table_59987.png} &
    \includegraphics[width = 0.032\linewidth]{icon/noun_kitchen_pot_3363643.png} &
    \includegraphics[width = 0.032\linewidth]{icon/noun_Door_1549119.png} &
    {\rotatebox{60}{\textbf{Avg.}}} \\
    \hline
    \multicolumn{22}{c}{\cellcolor[HTML]{FFC9C9}{\textbf{Success rate $\uparrow$}}} \\ 
    \hline
    ISS\cite{zhong2009intrinsic} & 0.43 & 0.44 & 0.52 & \textbf{0.99} & 0.73 & 0.85 & 0.45 & 0.61 & 0.55 & 0.68 & 0.63 & 0.68 & 0.48 & 0.90 & 0.67 & 0.59 & 0.78 & 0.25 & 0.78 & 0.64 & 0.64 \\ 
    
    SNAKE\cite{zhong2022snake} & 0.44 & 0.32 & 0.33 & 0.71 & 0.32 & 0.64 & 0.32 & 0.44 & 0.41 & 0.21 & 0.41 & 0.45 & 0.42 & 0.44 & 0.62 & 0.38 & 0.14 & 0.23 & 0.34 & 0.67 & 0.41 \\
    
    D3feat\cite{bai2020d3feat} & 0.55 & 0.57 & 0.80 & \textbf{0.99} & 0.43 & 0.60 & 0.52 & 0.73 & 0.34 & 0.67 & 0.62 & 0.56 & 0.67 & 0.26 & 0.61 & 0.09 & 0.41 & 0.55 & 0.88 & 0.76 & 0.53 \\ 
    
    INVERSE\cite{agrawal2016learning} & 0.43 & 0.68 & 0.72 & 0.55 & 0.63 & 0.89 & 0.78 & 0.65 & 0.61 & 0.83 & 0.68 & 0.67 & 0.59 & 0.80 & 0.73 & 0.58 & 0.83 & 0.67 & \textbf{1.00} & 0.68 & 0.73 \\
   
    UMPNet\cite{xu2022umpnet} & 0.67 & 0.78 & 0.90 & 0.73 & 0.68 & 0.86 & 0.90 & 0.58 & 0.63 & 0.79 & 0.75 & 0.68 & 0.89 & 0.86 & 0.76 & 0.62 & 0.80 & \textbf{0.68} & \textbf{1.00} & \textbf{0.79} & 0.79 \\

    Ours & \textbf{0.77} & \textbf{0.81} & \textbf{1.00} & 0.98 & \textbf{0.93} & \textbf{0.90} & \textbf{0.91} & \textbf{0.87} & \textbf{0.64} & \textbf{0.90} & \textbf{0.87} & \textbf{0.77} & \textbf{0.98} & \textbf{0.97} & \textbf{0.87} & \textbf{0.78} & \textbf{0.89} & 0.57 & 0.95 & 0.78 & \textbf{0.83} \\
    
    \multicolumn{22}{c}{\cellcolor[HTML]{FFC9C9}\textbf{Normalized distance to target $\downarrow$}} \\ \hline
    ISS\cite{zhong2009intrinsic} & 0.53 & 0.54 & 0.43 & \textbf{0.01} & 0.25 & 0.14 & 0.54 & 0.42 & 0.37 & 0.44 & 0.37 & 0.26 & 0.50 & 0.10 & 0.33 & 0.37 & 0.20 & 0.75 & 0.20 & 0.34 & 0.34 \\
    
    SNAKE\cite{zhong2022snake} & 0.52 & 0.62 & 0.63 & 0.28 & 0.67 & 0.33 & 0.67 & 0.54 & 0.58 & 0.78 & 0.56 & 0.54 & 0.53 & 0.55 & 0.35 & 0.61 & 0.86 & 0.77 & 0.66 & 0.31 & 0.58 \\
    
    D3feat\cite{bai2020d3feat} & 0.42 & 0.41 & 0.16 & \textbf{0.01} & 0.57 & 0.39 & 0.46 & 0.27 & 0.65 & 0.32 & 0.37 & 0.41 & 0.35 & 0.74 & 0.36 & 0.91 & 0.58 & 0.44 & 0.08 & 0.20 & 0.45 \\
    
    INVERSE\cite{agrawal2016learning} & 0.30 & 0.21 & 0.32 & 0.31 & 0.27 & 0.17 & 0.28 & \textbf{0.09} & \textbf{0.27} & \textbf{0.09} & 0.23 & 0.25 & 0.32 & 0.09 & 0.17 & 0.27 & 0.15 & \textbf{0.21} & \textbf{0.00} & 0.27 & 0.19 \\
    
    UMPNET\cite{xu2022umpnet} & \textbf{0.20} & 0.19 & 0.05 & 0.19 & 0.23 & 0.16 & 0.12 & 0.13 & 0.28 & 0.11 & 0.17 & 0.26 & 0.03 & 0.06 & 0.15 & \textbf{0.21} & 0.16 & 0.22 & \textbf{0.00} & \textbf{0.17} & \textbf{0.14} \\
    
    Ours & \textbf{0.20} & \textbf{0.09} & \textbf{0.00} & \textbf{0.01} & \textbf{0.05} & \textbf{0.08} & \textbf{0.07} & 0.10 & 0.35 & \textbf{0.09} & \textbf{0.11} & \textbf{0.19} & \textbf{0.02} & \textbf{0.02} & \textbf{0.08} & \textbf{0.21} & \textbf{0.11} & 0.42 & {0.05} & {0.21} & {0.15} \\ \hline
    
    \end{tabular}
}
\vspace{-2mm}
\end{table*}

\subsection{Goal Conditioned Manipulation}

The manipulation task is formulated in section~\ref{sec: manipulation}. We set the same initial and goal state for test objects, and the maximum action step as \cite{xu2022umpnet} used. We adopt our proposed manipulation strategy for other keypoint-based methods. \\
%
\textbf{Metrics}
Following UMPNet, we exploit 1) \textit{normalized distance to target state} and 2) \textit{success rate} as manipulation metrics. The first metric means the distance between the end and target state divided by the distance between the initial and target state, denoted as $d$. A success means $d \textless 0.1$. \\
\textbf{Evaluation Results}
We report the results in Tab.~\ref{tab:manipulation_results}. 
Our method surpasses other keypoint-based baselines on both seen and unseen categories, demonstrating the better performance of our predicted keypoints. Our method also outperforms manipulation-centric methods Inverse \cite{agrawal2016learning} and UMPNet \cite{xu2022umpnet} in most categories. 
It is noteworthy that, although UMPNET produces comparable results, it necessitates prohibitively costly trial-and-error simulations for
pixel-wise affordance learning. The training time of UMPNet is 5-7 times of ours when using the same hardware, which can be found in the supplementary. Notably, they also need color and surface normal for each point. It is suggested that the learning formulation of the 3D Implicit Transporter is efficient and the manipulation strategy based on correspondent keypoints is also effective. 
We provide the qualitative results of our method in Fig.~\ref{fig:goal conditioned manipulation}. With the correspondent keypoints, we can compute the axis-angle and translation direction, which direct us to generate a future action to interact. In contrast to moving an end-effector a fixed length every time as UMPNet used, we dynamically adjust the moving distance according to the articulation flow $\boldsymbol{\rm F}_{c,g}^{s}$ between the current and goal state. 

\subsection{Real-World Experiments}
\begin{figure}[h]
    \centering
    \vskip -0.15in
    \includegraphics[width=0.45\textwidth]{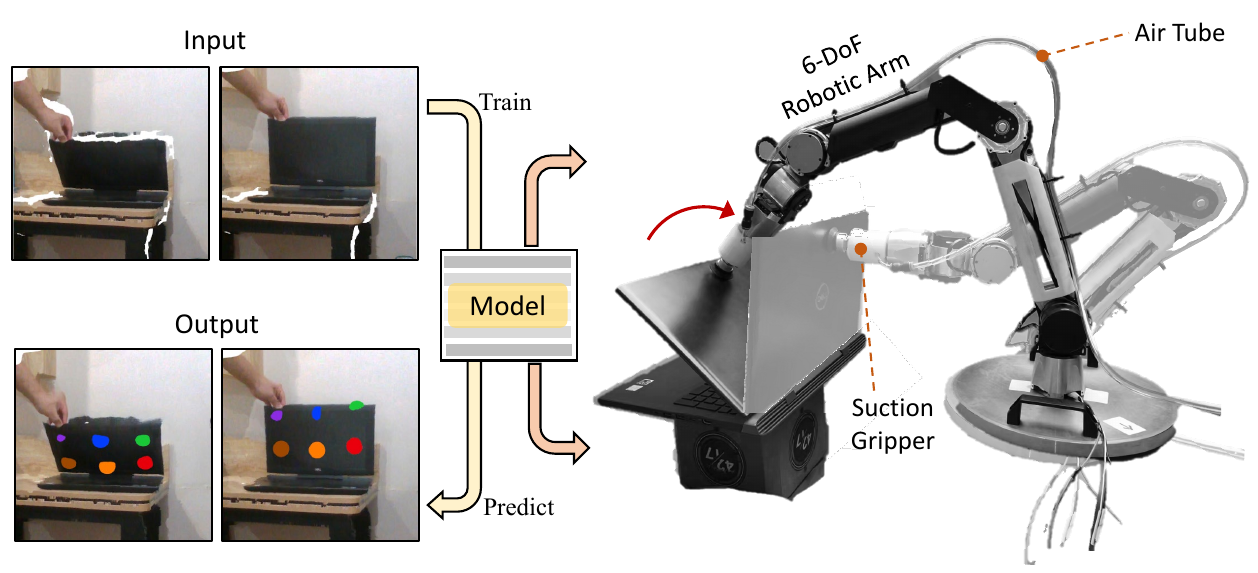}
    \vspace{-1.0mm}
    \caption{Real-world experiment. We train the 3D Transporter on the real data and transfer the model to a real robotic application.
    }
    \label{fig:real-world}
    \vskip -0.2in
\end{figure}

Finally, to examine the effectiveness of our method, we conduct a real-world experiment. We design a robot system that consists of a 6-DoF robotic arm, a pneumatic suction gripper as the end-effector, and a calibrated RGB-D camera, as visualized in Fig.~\ref{fig:real-world}. We choose a laptop as the manipulation object. Results on more real-world objects can be found in the supplementary. For training data generation, we interact with the laptop to change articulation states under different camera view-points. In contrast to synthetic data, real data contains human motion which encourages keypoints to locate on the human body according to the principle of the Transporter. Therefore, we manually label an bounding box in the first frame of each video to coarsely filter points of the laptop in all frames.
Fig.~\ref{fig:real-world} illustrates that our method could find 
temporally consistent keypoints and the proper suction positions and action directions. 

\section{Conclusions}
Our work presents 3D Implicit Transporter, a self-supervised method to discover temporally correspondent 3D keypoints from point cloud sequences. We introduce three novel components that extend the 2D Transporter to the 3D domain. 
Extensive evaluations show that our keypoints are temporally consistent and generalizable to unseen object categories. Moreover, we develop a manipulation policy for downstream tasks that utilizes the Transporter keypoints and demonstrate that they are suitable for 3D manipulation tasks.

\textbf{Acknowledgments.}
This research was sponsored by Baidu Inc. through Apollo-AIR Joint Research Center. It was also supported by National Natural Science Foundation of China (62006244), Young Elite Scientist Sponsorship Program of China (and Beijing) Association for Science and Technology (YESS20200140, BYESS2021178).


{\small
\bibliographystyle{ieee_fullname}
\bibliography{egbib}

\begin{thebibliography}{10}\itemsep=-1pt

\bibitem{abdul2022learning}
Hameed Abdul-Rashid, Miles Freeman, Ben Abbatematteo, George Konidaris, and
  Daniel Ritchie.
\newblock Learning to infer kinematic hierarchies for novel object instances.
\newblock In {\em ICRA}, 2022.

\bibitem{agrawal2016learning}
Pulkit Agrawal, Ashvin~V Nair, Pieter Abbeel, Jitendra Malik, and Sergey
  Levine.
\newblock Learning to poke by poking: Experiential learning of intuitive
  physics.
\newblock In {\em NeurIPS}, 2016.

\bibitem{ang2005pid}
Kiam~Heong Ang, Gregory Chong, and Yun Li.
\newblock Pid control system analysis, design, and technology.
\newblock {\em IEEE transactions on control systems technology},
  13(4):559--576, 2005.

\bibitem{bai2020d3feat}
Xuyang Bai, Zixin Luo, Lei Zhou, Hongbo Fu, Long Quan, and Chiew-Lan Tai.
\newblock D3feat: Joint learning of dense detection and description of 3d local
  features.
\newblock In {\em CVPR}, 2020.

\bibitem{baker2011database}
Simon Baker, Daniel Scharstein, JP Lewis, Stefan Roth, Michael~J Black, and
  Richard Szeliski.
\newblock A database and evaluation methodology for optical flow.
\newblock {\em IJCV}, 92(1):1--31, 2011.

\bibitem{barron1994performance}
John~L Barron, David~J Fleet, and Steven~S Beauchemin.
\newblock Performance of optical flow techniques.
\newblock {\em IJCV}, 12(1):43--77, 1994.

\bibitem{bay2008speeded}
Herbert Bay, Andreas Ess, Tinne Tuytelaars, and Luc Van~Gool.
\newblock Speeded-up robust features (surf).
\newblock {\em CVIU}, 110(3):346--359, 2008.

\bibitem{besl1992method}
Paul~J Besl and Neil~D McKay.
\newblock Method for registration of 3-d shapes.
\newblock In {\em Sensor fusion IV: control paradigms and data structures},
  volume 1611, pages 586--606. Spie, 1992.

\bibitem{butler2012naturalistic}
Daniel~J Butler, Jonas Wulff, Garrett~B Stanley, and Michael~J Black.
\newblock A naturalistic open source movie for optical flow evaluation.
\newblock In {\em ECCV}, 2012.

\bibitem{calonder2010brief}
Michael Calonder, Vincent Lepetit, Christoph Strecha, and Pascal Fua.
\newblock Brief: Binary robust independent elementary features.
\newblock In {\em ECCV}, 2010.

\bibitem{chen2021unsupervised}
Boyuan Chen, Pieter Abbeel, and Deepak Pathak.
\newblock Unsupervised learning of visual 3d keypoints for control.
\newblock In {\em ICML}, 2021.

\bibitem{cciccek20163d}
{\"O}zg{\"u}n {\c{C}}i{\c{c}}ek, Ahmed Abdulkadir, Soeren~S Lienkamp, Thomas
  Brox, and Olaf Ronneberger.
\newblock 3d u-net: learning dense volumetric segmentation from sparse
  annotation.
\newblock In {\em MICCAI}, 2016.

\bibitem{dosovitskiy2015flownet}
Alexey Dosovitskiy, Philipp Fischer, Eddy Ilg, Philip Hausser, Caner Hazirbas,
  Vladimir Golkov, Patrick Van Der~Smagt, Daniel Cremers, and Thomas Brox.
\newblock Flownet: Learning optical flow with convolutional networks.
\newblock In {\em ICCV}, 2015.

\bibitem{EisnerZhang2022FLOW}
Ben Eisner*, Harry Zhang*, and David Held.
\newblock Flowbot3d: Learning 3d articulation flow to manipulate articulated
  objects.
\newblock In {\em RSS}, 2022.

\bibitem{fan2017point}
Haoqiang Fan, Hao Su, and Leonidas~J Guibas.
\newblock A point set generation network for 3d object reconstruction from a
  single image.
\newblock In {\em CVPR}, 2017.

\bibitem{florence2019self}
Peter Florence, Lucas Manuelli, and Russ Tedrake.
\newblock Self-supervised correspondence in visuomotor policy learning.
\newblock {\em IEEE Robotics and Automation Letters}, 5(2):492--499, 2019.

\bibitem{gopalakrishnan2020unsupervised}
Anand Gopalakrishnan, Sjoerd van Steenkiste, and J{\"u}rgen Schmidhuber.
\newblock Unsupervised object keypoint learning using local spatial
  predictability.
\newblock {\em arXiv preprint arXiv:2011.12930}, 2020.

\bibitem{haque2016towards}
Albert Haque, Boya Peng, Zelun Luo, Alexandre Alahi, Serena Yeung, and Li
  Fei-Fei.
\newblock Towards viewpoint invariant 3d human pose estimation.
\newblock In {\em ECCV}, 2016.

\bibitem{horn1981determining}
Berthold~KP Horn and Brian~G Schunck.
\newblock Determining optical flow.
\newblock {\em Artificial intelligence}, 17(1-3):185--203, 1981.

\bibitem{jiang2022ditto}
Zhenyu Jiang, Cheng-Chun Hsu, and Yuke Zhu.
\newblock Ditto: Building digital twins of articulated objects from
  interaction.
\newblock {\em arXiv preprint arXiv:2202.08227}, 2022.

\bibitem{jiang2021synergies}
Zhenyu Jiang, Yifeng Zhu, Maxwell Svetlik, Kuan Fang, and Yuke Zhu.
\newblock Synergies between affordance and geometry: 6-dof grasp detection via
  implicit representations.
\newblock {\em arXiv preprint arXiv:2104.01542}, 2021.

\bibitem{katz2008manipulating}
Dov Katz and Oliver Brock.
\newblock Manipulating articulated objects with interactive perception.
\newblock In {\em ICRA}, 2008.

\bibitem{kingma2017adam}
Diederik~P. Kingma and Jimmy Ba.
\newblock Adam: A method for stochastic optimization, 2017.

\bibitem{kulkarni2019unsupervised}
Tejas~D Kulkarni, Ankush Gupta, Catalin Ionescu, Sebastian Borgeaud, Malcolm
  Reynolds, Andrew Zisserman, and Volodymyr Mnih.
\newblock Unsupervised learning of object keypoints for perception and control.
\newblock In {\em NeurIPS}, 2019.

\bibitem{lee2005mesh}
Chang~Ha Lee, Amitabh Varshney, and David~W Jacobs.
\newblock Mesh saliency.
\newblock In {\em SIGGRAPH}. 2005.

\bibitem{li2019usip}
Jiaxin Li and Gim~Hee Lee.
\newblock Usip: Unsupervised stable interest point detection from 3d point
  clouds.
\newblock In {\em CVPR}, 2019.

\bibitem{Li2023LODELC}
Pengfei Li, Ruowen Zhao, Yongliang Shi, Hao Zhao, Jirui Yuan, Guyue Zhou, and
  Ya-Qin Zhang.
\newblock Lode: Locally conditioned eikonal implicit scene completion from
  sparse lidar.
\newblock {\em 2023 IEEE International Conference on Robotics and Automation
  (ICRA)}, pages 8269--8276, 2023.

\bibitem{li2020category}
Xiaolong Li, He Wang, Li Yi, Leonidas~J Guibas, A~Lynn Abbott, and Shuran Song.
\newblock Category-level articulated object pose estimation.
\newblock In {\em CVPR}, 2020.

\bibitem{li2022lepard}
Yang Li and Tatsuya Harada.
\newblock Lepard: Learning partial point cloud matching in rigid and deformable
  scenes.
\newblock In {\em Proceedings of the IEEE/CVF conference on computer vision and
  pattern recognition}, pages 5554--5564, 2022.

\bibitem{liu2020neural}
Lingjie Liu, Jiatao Gu, Kyaw Zaw~Lin, Tat-Seng Chua, and Christian Theobalt.
\newblock Neural sparse voxel fields.
\newblock In {\em NeurIPS}, 2020.

\bibitem{lowe2004distinctive}
David~G Lowe.
\newblock Distinctive image features from scale-invariant keypoints.
\newblock {\em IJCV}, 60(2):91--110, 2004.

\bibitem{ma2022learning}
Xiao Ma, David Hsu, and Wee~Sun Lee.
\newblock Learning latent graph dynamics for visual manipulation of deformable
  objects.
\newblock In {\em ICRA}.

\bibitem{manuelli2022kpam}
Lucas Manuelli, Wei Gao, Peter Florence, and Russ Tedrake.
\newblock kpam: Keypoint affordances for category-level robotic manipulation.
\newblock In {\em Robotics Research: The 19th International Symposium ISRR}.

\bibitem{mildenhall2020nerf}
Ben Mildenhall, Pratul~P Srinivasan, Matthew Tancik, Jonathan~T Barron, Ravi
  Ramamoorthi, and Ren Ng.
\newblock Nerf: Representing scenes as neural radiance fields for view
  synthesis.
\newblock In {\em ECCV}, 2020.

\bibitem{minderer2019unsupervised}
Matthias Minderer, Chen Sun, Ruben Villegas, Forrester Cole, Kevin~P Murphy,
  and Honglak Lee.
\newblock Unsupervised learning of object structure and dynamics from videos.
\newblock In {\em NeurIPS}, 2019.

\bibitem{mo2021where2act}
Kaichun Mo, Leonidas~J Guibas, Mustafa Mukadam, Abhinav Gupta, and Shubham
  Tulsiani.
\newblock Where2act: From pixels to actions for articulated 3d objects.
\newblock In {\em CVPR}, 2021.

\bibitem{mu2021sdf}
Jiteng Mu, Weichao Qiu, Adam Kortylewski, Alan Yuille, Nuno Vasconcelos, and
  Xiaolong Wang.
\newblock A-sdf: Learning disentangled signed distance functions for
  articulated shape representation.
\newblock In {\em CVPR}, 2021.

\bibitem{mur2015orb}
Raul Mur-Artal, Jose Maria~Martinez Montiel, and Juan~D Tardos.
\newblock Orb-slam: a versatile and accurate monocular slam system.
\newblock {\em IEEE transactions on robotics}, 31(5):1147--1163, 2015.

\bibitem{nie2022structure}
Neil Nie, Samir~Yitzhak Gadre, Kiana Ehsani, and Shuran Song.
\newblock Structure from action: Learning interactions for articulated object
  3d structure discovery.
\newblock {\em arXiv preprint arXiv:2207.08997}, 2022.

\bibitem{park2019deepsdf}
Jeong~Joon Park, Peter Florence, Julian Straub, Richard Newcombe, and Steven
  Lovegrove.
\newblock Deepsdf: Learning continuous signed distance functions for shape
  representation.
\newblock In {\em CVPR}, 2019.

\bibitem{pytorch}
Adam Paszke, Sam Gross, Francisco Massa, Adam Lerer, James Bradbury, Gregory
  Chanan, Trevor Killeen, Zeming Lin, Natalia Gimelshein, Luca Antiga, Alban
  Desmaison, Andreas Kopf, Edward Yang, Zachary DeVito, Martin Raison, Alykhan
  Tejani, Sasank Chilamkurthy, Benoit Steiner, Lu Fang, Junjie Bai, and Soumith
  Chintala.
\newblock Pytorch: An imperative style, high-performance deep learning library.
\newblock In {\em Advances in Neural Information Processing Systems 32}, pages
  8024--8035. Curran Associates, Inc., 2019.

\bibitem{patel2023animal}
Mahir Patel, Yiwen Gu, Lucas~C Carstensen, Michael~E Hasselmo, and Margrit
  Betke.
\newblock Animal pose tracking: 3d multimodal dataset and token-based pose
  optimization.
\newblock {\em International Journal of Computer Vision}, 131(2):514--530,
  2023.

\bibitem{convonet}
Songyou Peng, Michael Niemeyer, Lars Mescheder, Marc Pollefeys, and Andreas
  Geiger.
\newblock Convolutional occupancy networks.
\newblock In {\em ECCV}, 2020.

\bibitem{qi2017pointnetplusplus}
Charles~Ruizhongtai Qi, Li Yi, Hao Su, and Leonidas~J Guibas.
\newblock Pointnet++: Deep hierarchical feature learning on point sets in a
  metric space.
\newblock In {\em NeurIPS}, 2017.

\bibitem{qin2020keto}
Zengyi Qin, Kuan Fang, Yuke Zhu, Li Fei-Fei, and Silvio Savarese.
\newblock Keto: Learning keypoint representations for tool manipulation.
\newblock In {\em ICRA}, 2020.

\bibitem{rahmani2014hopc}
Hossein Rahmani, Arif Mahmood, Q Du~Huynh, and Ajmal Mian.
\newblock Hopc: Histogram of oriented principal components of 3d pointclouds
  for action recognition.
\newblock In {\em ECCV}, 2014.

\bibitem{rublee2011orb}
Ethan Rublee, Vincent Rabaud, Kurt Konolige, and Gary Bradski.
\newblock Orb: An efficient alternative to sift or surf.
\newblock In {\em ICCV}, 2011.

\bibitem{rusu2009fast}
Radu~Bogdan Rusu, Nico Blodow, and Michael Beetz.
\newblock Fast point feature histograms (fpfh) for 3d registration.
\newblock In {\em ICRA}, 2009.

\bibitem{superglue}
Paul-Edouard Sarlin, Daniel DeTone, Tomasz Malisiewicz, and Andrew Rabinovich.
\newblock Superglue: Learning feature matching with graph neural networks.
\newblock In {\em CVPR}, 2020.

\bibitem{schwarz2020graf}
Katja Schwarz, Yiyi Liao, Michael Niemeyer, and Andreas Geiger.
\newblock Graf: Generative radiance fields for 3d-aware image synthesis.
\newblock In {\em NeurIPS}, 2020.

\bibitem{simeonov2022neural}
Anthony Simeonov, Yilun Du, Andrea Tagliasacchi, Joshua~B Tenenbaum, Alberto
  Rodriguez, Pulkit Agrawal, and Vincent Sitzmann.
\newblock Neural descriptor fields: Se (3)-equivariant object representations
  for manipulation.
\newblock In {\em ICRA}, 2022.

\bibitem{sitzmann2020implicit}
Vincent Sitzmann, Julien Martel, Alexander Bergman, David Lindell, and Gordon
  Wetzstein.
\newblock Implicit neural representations with periodic activation functions.
\newblock In {\em NeurIPS}, 2020.

\bibitem{sivic2003video}
Josef Sivic and Andrew Zisserman.
\newblock Video google: A text retrieval approach to object matching in videos.
\newblock In {\em ICCV}, 2003.

\bibitem{spelke1990principles}
Elizabeth~S Spelke.
\newblock Principles of object perception.
\newblock {\em Cognitive science}, 14(1):29--56, 1990.

\bibitem{sun2022disentangling}
Deqing Sun, Charles Herrmann, Fitsum Reda, Michael Rubinstein, David~J Fleet,
  and William~T Freeman.
\newblock Disentangling architecture and training for optical flow.
\newblock In {\em ECCV}, 2022.

\bibitem{sun2010secrets}
Deqing Sun, Stefan Roth, and Michael~J Black.
\newblock Secrets of optical flow estimation and their principles.
\newblock In {\em CVPR}, 2010.

\bibitem{sun2018pwc}
Deqing Sun, Xiaodong Yang, Ming-Yu Liu, and Jan Kautz.
\newblock Pwc-net: Cnns for optical flow using pyramid, warping, and cost
  volume.
\newblock In {\em CVPR}, 2018.

\bibitem{sun2022self}
Jennifer~J Sun, Serim Ryou, Roni~H Goldshmid, Brandon Weissbourd, John~O
  Dabiri, David~J Anderson, Ann Kennedy, Yisong Yue, and Pietro Perona.
\newblock Self-supervised keypoint discovery in behavioral videos.
\newblock In {\em CVPR}, 2022.

\bibitem{suwajanakorn2018discovery}
Supasorn Suwajanakorn, Noah Snavely, Jonathan~J Tompson, and Mohammad Norouzi.
\newblock Discovery of latent 3d keypoints via end-to-end geometric reasoning.
\newblock In {\em NeurIPS}, 2018.

\bibitem{tchapmi2019topnet}
Lyne~P Tchapmi, Vineet Kosaraju, Hamid Rezatofighi, Ian Reid, and Silvio
  Savarese.
\newblock Topnet: Structural point cloud decoder.
\newblock In {\em CVPR}, 2019.

\bibitem{vaswani2017attention}
Ashish Vaswani, Noam Shazeer, Niki Parmar, Jakob Uszkoreit, Llion Jones,
  Aidan~N Gomez, {\L}ukasz Kaiser, and Illia Polosukhin.
\newblock Attention is all you need.
\newblock In {\em NeurIPS}, 2017.

\bibitem{wang2019densefusion}
Chen Wang, Danfei Xu, Yuke Zhu, Roberto Mart{\'\i}n-Mart{\'\i}n, Cewu Lu, Li
  Fei-Fei, and Silvio Savarese.
\newblock Densefusion: 6d object pose estimation by iterative dense fusion.
\newblock In {\em CVPR}.

\bibitem{wang2021unsupervised}
Xudong Wang, Long Lian, and Stella~X Yu.
\newblock Unsupervised visual attention and invariance for reinforcement
  learning.
\newblock In {\em CVPR}, 2021.

\bibitem{wang2019shape2motion}
Xiaogang Wang, Bin Zhou, Yahao Shi, Xiaowu Chen, Qinping Zhao, and Kai Xu.
\newblock Shape2motion: Joint analysis of motion parts and attributes from 3d
  shapes.
\newblock In {\em CVPR}, 2019.

\bibitem{wang2022adaafford}
Yian Wang, Ruihai Wu, Kaichun Mo, Jiaqi Ke, Qingnan Fan, Leonidas~J Guibas, and
  Hao Dong.
\newblock Adaafford: Learning to adapt manipulation affordance for 3d
  articulated objects via few-shot interactions.
\newblock In {\em ECCV}, 2022.

\bibitem{weng2021captra}
Yijia Weng, He Wang, Qiang Zhou, Yuzhe Qin, Yueqi Duan, Qingnan Fan, Baoquan
  Chen, Hao Su, and Leonidas~J Guibas.
\newblock Captra: Category-level pose tracking for rigid and articulated
  objects from point clouds.
\newblock In {\em CVPR}, 2021.

\bibitem{wu2021vat}
Ruihai Wu, Yan Zhao, Kaichun Mo, Zizheng Guo, Yian Wang, Tianhao Wu, Qingnan
  Fan, Xuelin Chen, Leonidas Guibas, and Hao Dong.
\newblock Vat-mart: Learning visual action trajectory proposals for
  manipulating 3d articulated objects.
\newblock {\em arXiv preprint arXiv:2106.14440}, 2021.

\bibitem{xiang2020sapien}
Fanbo Xiang, Yuzhe Qin, Kaichun Mo, Yikuan Xia, Hao Zhu, Fangchen Liu, Minghua
  Liu, Hanxiao Jiang, Yifu Yuan, He Wang, et~al.
\newblock Sapien: A simulated part-based interactive environment.
\newblock In {\em CVPR}, 2020.

\bibitem{xiang2017posecnn}
Yu Xiang, Tanner Schmidt, Venkatraman Narayanan, and Dieter Fox.
\newblock Posecnn: A convolutional neural network for 6d object pose estimation
  in cluttered scenes.
\newblock {\em arXiv preprint arXiv:1711.00199}, 2017.

\bibitem{xu2022umpnet}
Zhenjia Xu, He Zhanpeng, and Shuran Song.
\newblock Umpnet: Universal manipulation policy network for articulated
  objects.
\newblock {\em RA-L}, 7:2447--2454, 2022.

\bibitem{yi2018deep}
Li Yi, Haibin Huang, Difan Liu, Evangelos Kalogerakis, Hao Su, and Leonidas
  Guibas.
\newblock Deep part induction from articulated object pairs.
\newblock {\em arXiv preprint arXiv:1809.07417}, 2018.

\bibitem{you2022ukpgan}
Yang You, Wenhai Liu, Yanjie Ze, Yong-Lu Li, Weiming Wang, and Cewu Lu.
\newblock Ukpgan: A general self-supervised keypoint detector.
\newblock In {\em CVPR}, 2022.

\bibitem{Zeng20173DMatchLL}
Andy Zeng, Shuran Song, Matthias Nie{\ss}ner, Matthew Fisher, Jianxiong Xiao,
  and Thomas~A. Funkhouser.
\newblock 3dmatch: Learning local geometric descriptors from rgb-d
  reconstructions.
\newblock In {\em CVPR}, 2017.

\bibitem{zeng2021corrnet3d}
Yiming Zeng, Yue Qian, Zhiyu Zhu, Junhui Hou, Hui Yuan, and Ying He.
\newblock Corrnet3d: Unsupervised end-to-end learning of dense correspondence
  for 3d point clouds.
\newblock In {\em Proceedings of the IEEE/CVF Conference on Computer Vision and
  Pattern Recognition}, pages 6052--6061, 2021.

\bibitem{zhao2020fine}
Jian Zhao, Jianshu Li, Hengzhu Liu, Shuicheng Yan, and Jiashi Feng.
\newblock Fine-grained multi-human parsing.
\newblock {\em International Journal of Computer Vision}, 2020.

\bibitem{zhong2022sim2real}
Chengliang Zhong, Chao Yang, Fuchun Sun, Jinshan Qi, Xiaodong Mu, Huaping Liu,
  and Wenbing Huang.
\newblock Sim2real object-centric keypoint detection and description.
\newblock In {\em AAAI}, 2022.

\bibitem{zhong2022snake}
Chengliang Zhong, Peixing You, Xiaoxue Chen, Hao Zhao, Fuchun Sun, Guyue Zhou,
  Xiaodong Mu, Chuang Gan, and Wenbing Huang.
\newblock Snake: Shape-aware neural 3d keypoint field.
\newblock {\em Advances in Neural Information Processing Systems},
  35:7052--7064, 2022.

\bibitem{zhong2009intrinsic}
Yu Zhong.
\newblock Intrinsic shape signatures: A shape descriptor for 3d object
  recognition.
\newblock In {\em ICCV Workshops}, 2009.

\bibitem{zhou2016fast}
Qian-Yi Zhou, Jaesik Park, and Vladlen Koltun.
\newblock Fast global registration.
\newblock In {\em ECCV}, 2016.

\bibitem{zhu1998filters}
Song~Chun Zhu, Yingnian Wu, and David Mumford.
\newblock Filters, random fields and maximum entropy (frame): Towards a unified
  theory for texture modeling.
\newblock {\em IJCV}, 27(2):107--126, 1998.

\end{thebibliography}
}

\clearpage
\appendix

{\Large \textbf{Appendix}}
\section{Implementation Details}

We implement our models in PyTorch~\cite{pytorch} with the Adam~\cite{kingma2017adam} optimizer and a mini-batch size of 10 on 4 NVIDIA A100 GPUs for 45 epochs. 
A learning rate of $10^{-4}$ is used for the first 30 epochs, which is dropped ten times for the remainder.
To increase data diversity, we perform random rigid transformation and Gaussian noise for input point clouds. 
Perception and manipulation hyper-parameters are provided in Tab.~\ref{tab:hyper}. 

\begin{table}[htbp]
  \centering
  \label{tab:hyper}
  \caption{Perception and manipulation hyper-parameters.}
  \resizebox{0.99\columnwidth}{!}{
    \begin{tabular}{cc|ccc|c|cc}
    \toprule
    $N_1$     & $N_2$    & $C_1/C_2/C_3/C_5/C_e$  & $C_4$   & $C_h/C_w/C_d$ & $\lambda_1$/$\lambda_2$ & $\theta_{thr}$ & $\lambda$ \\
    \midrule
    5000  & 128   & 32   & 256   & 64 & 1 & 0.1   & 8 \\
    \bottomrule
    \end{tabular}%
}
\end{table}%

\begin{figure*}[ht]
  \centering
  \includegraphics[width=0.95\linewidth]{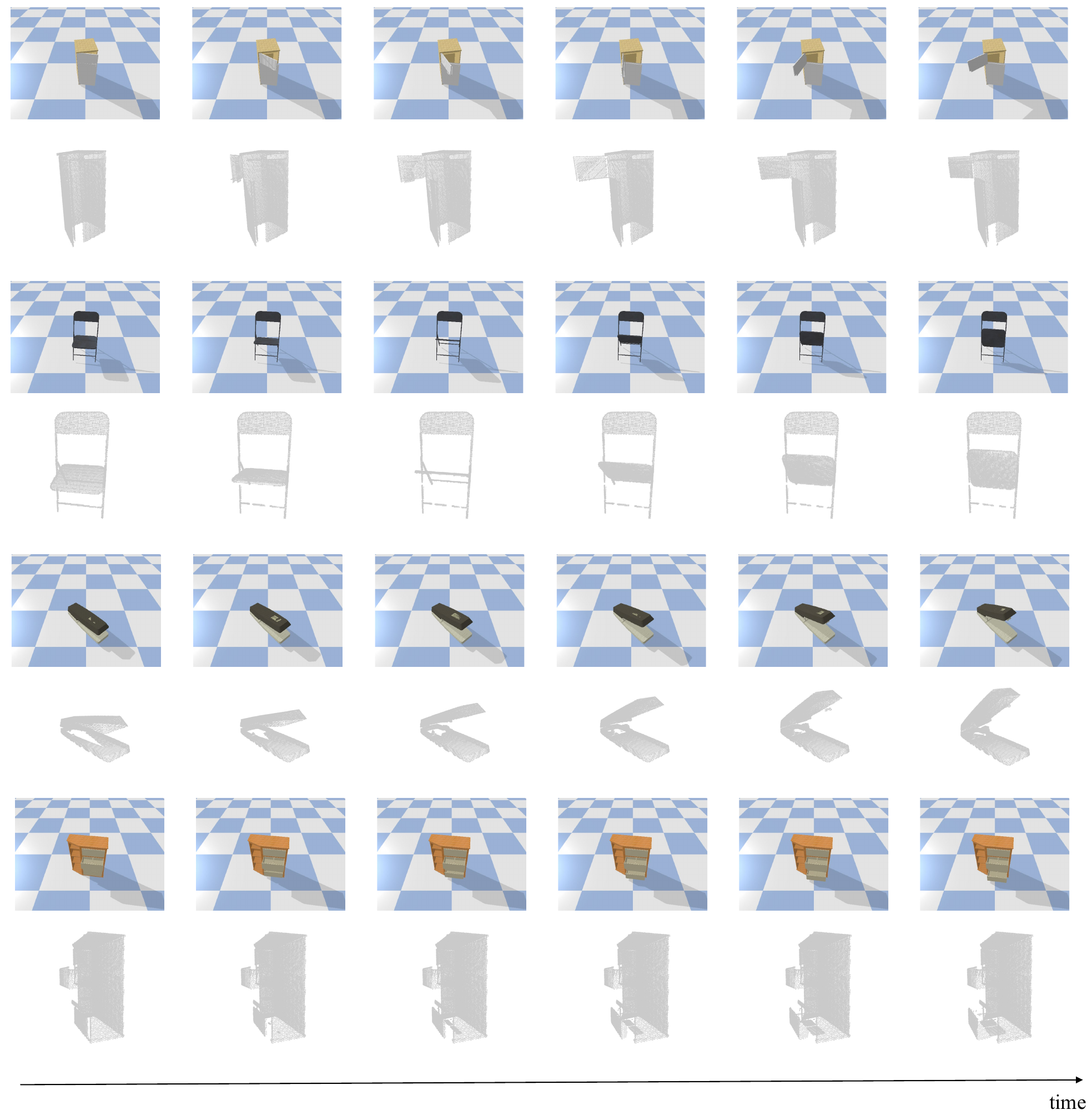}
  \caption{Examples of sequences of training data.}
  \label{fig:training}
\end{figure*}

Fig.~\ref{fig:training} depicts a series of training examples comprising rendered RGB images, accompanied by corresponding point clouds, for an articulated object with motion in its constituent parts. It is pertinent to note that solely the rendered point clouds are utilized for training and testing purposes.


\section{Ablation Study}
\paragraph{Keypoint Parameters}
Tab.~\ref{tab:kp} provides the quantitative results on keypoint parameters. Increasing keypoint number $m$ and Gaussian variance $\sigma$ can transport more features from target to source so that the reconstruction performance improves. However, few keypoints are enough for objects with relatively small mobile parts to transport core features. In this case, the redundant keypoints may scatter on stationary parts, which could harm pose estimation on mobile parts. For our training data, $m=6$ and $\sigma=0.15$ are the best choice.

\begin{table}[t]
\caption{Ablation study of parameters of keypoint network.}
\label{tab:kp}
\begin{center}
\resizebox{0.85\columnwidth}{!}{
\begin{tabular}{cc|ccc}
\hline
\textbf{$m$} & \textbf{$\sigma$} & \textbf{RR $\uparrow$} & \textbf{ACKD $\downarrow$} & \textbf{ADD $\downarrow$} \\ \hline
\rowcolor[HTML]{EFEFEF}
5 & 0.15 & 0.623 & 0.130 & 0.129 \\
6 & 0.10 & 0.604 & 0.136 & 0.130 \\
\rowcolor[HTML]{EFEFEF}
6 & 0.15 & 0.611 & \textbf{0.120} & \textbf{0.109} \\
6 & 0.20 & \textbf{0.601} & 0.128 & 0.122 \\
\rowcolor[HTML]{EFEFEF}
7 & 0.15 & 0.606 & 0.125 & 0.113 \\
8 & 0.15 & 0.551 & 0.141 & 0.130 \\ \hline
\end{tabular}
}
\end{center}
\end{table}

\paragraph{Volume Size}
We have conducted an ablation study of the impact of the volume size. The results are reported in Tab.~\ref{tab:volume_size}. The higher volumetric resolution of feature grids improves keypoint detection performance but increases the computation cost. Therefore, we choose the voxel size of 64 to balance the memory cost and perception performance.

\begin{table}[t]
  \centering
  \caption{The impact of volume size.}
  \resizebox{0.85\columnwidth}{!}{
    \begin{tabular}{c|ccc}
    \hline
    $C_h/C_w/C_d$ & \textbf{RR $\uparrow$} & \textbf{ACKD $\downarrow$} & \textbf{ADD $\downarrow$}  \\
    \hline
    \rowcolor[HTML]{EFEFEF}
    16    & 0.567 & 0.147 & 0.153  \\
    32    & \textbf{0.642} & 0.125 & 0.123 \\
    \rowcolor[HTML]{EFEFEF}
    64    & 0.611 & \textbf{0.120} & \textbf{0.109}   \\
    \hline
    \end{tabular}}%
    \label{tab:volume_size}
\end{table}%


\section{Formulation of the Additional Loss }
As discussed in the main paper, we incorporate an additional loss term, $\mathcal{L}_{\rm occ\_s}$, to facilitate the source frame reconstruction for better perception results. 

Via the volume features $\mit \Phi(\boldsymbol{{\rm o}}_{s})$ of the source frame and the corresponding query set, the geometry decoder is required to predict the occupancy of the source frame, which is written as:
\begin{align}
    \label{equ:occp_s}
    {\mit \Omega }(\boldsymbol{{\rm q}}_e, \mit \Phi_{\boldsymbol{{\rm q}}}(\boldsymbol{{\rm o}}_s)) \rightarrow {\rm Prob}(\boldsymbol{{\rm q}}|\boldsymbol{{\rm o}}_s)
\end{align}

Then, we use the binary cross-entropy loss to assess the dissimilarity between the decoded and the priori specified occupancy values by:
\begin{align}
    \label{equ:occp_loss_s}
    \mathcal{L}_{\rm occ\_s}=\frac{1}{|Q|}\sum_{\boldsymbol{{\rm q}}\in Q}l_{\rm BCE} \big({\rm Prob}(\boldsymbol{{\rm q}}|\boldsymbol{{\rm o}}_s), {\rm Prob}^{\rm gt}(\boldsymbol{{\rm q}}|\boldsymbol{{\rm o}}_s)\big)
\end{align}

\section{Training Efficiency}
\begin{figure}[ht]
  \centering
  \includegraphics[width=0.99\linewidth]{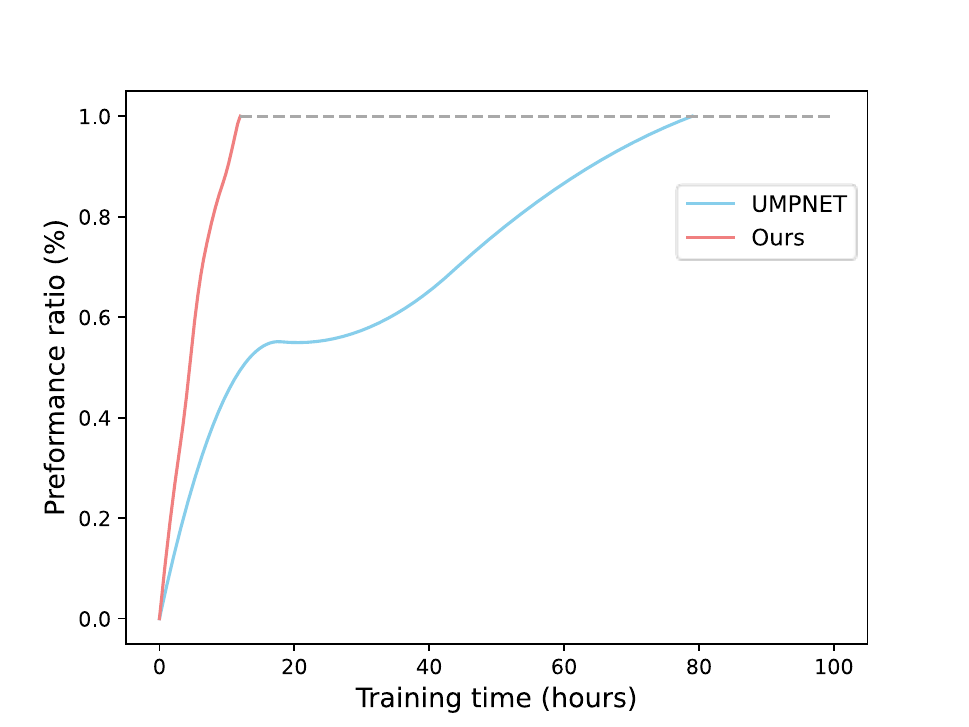}
  \caption{The training time (in hours) required to achieve the best performance of both UMPNET and ours.}
  \label{fig:kp_per}
\end{figure}

We compare our training time cost with UMPNET. Fig.~\ref{fig:kp_per} presents a comparative analysis of the training time cost required to achieve the best performance of both UMPNET and our proposed model, utilizing the same hardware (an Nvidia A100 GPU). The presented results demonstrate the superior efficiency of our 3D Implicit Transporter. Our belief is that this can be attributed to the more efficient nature of sparse keypoint learning as opposed to dense affordance prediction.

\begin{figure}[ht]
    \centering
    \includegraphics[width=0.99\linewidth]{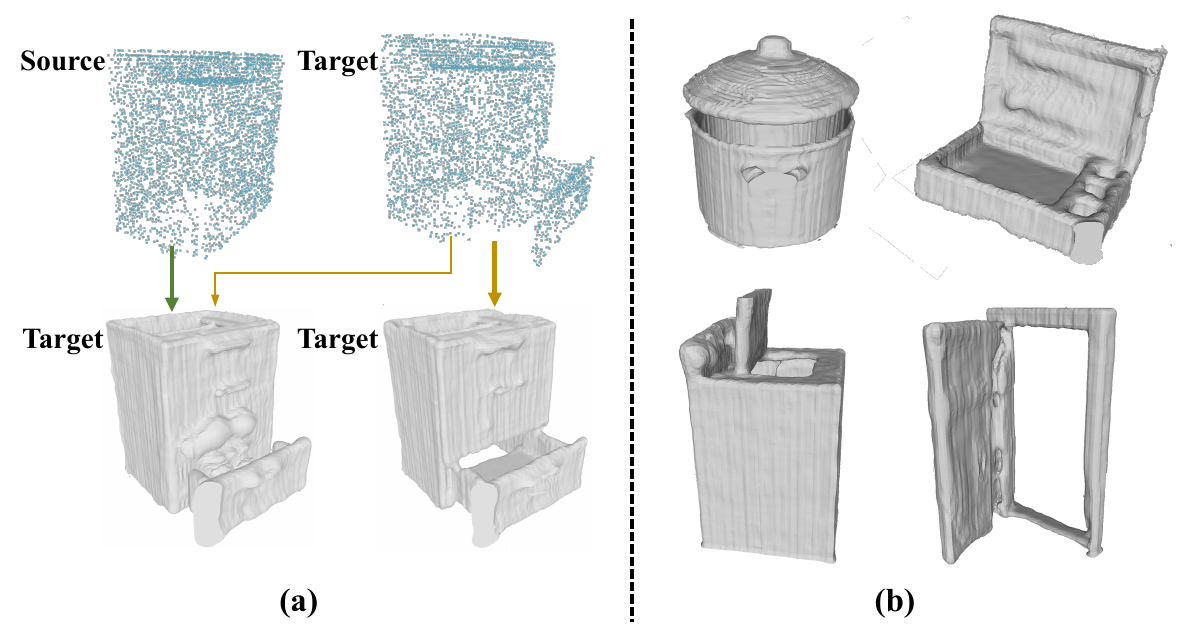}
    \caption{Surface shape reconstruction. (a) Target shape reconstruction from transported source features and target features, respectively. (b) Reconstruction results of unseen categories.}
    \label{fig:trans_unseen}
\end{figure}

\section{Qualitative Results}
\begin{figure*}[ht]
  \centering
  \includegraphics[width=0.79\linewidth]{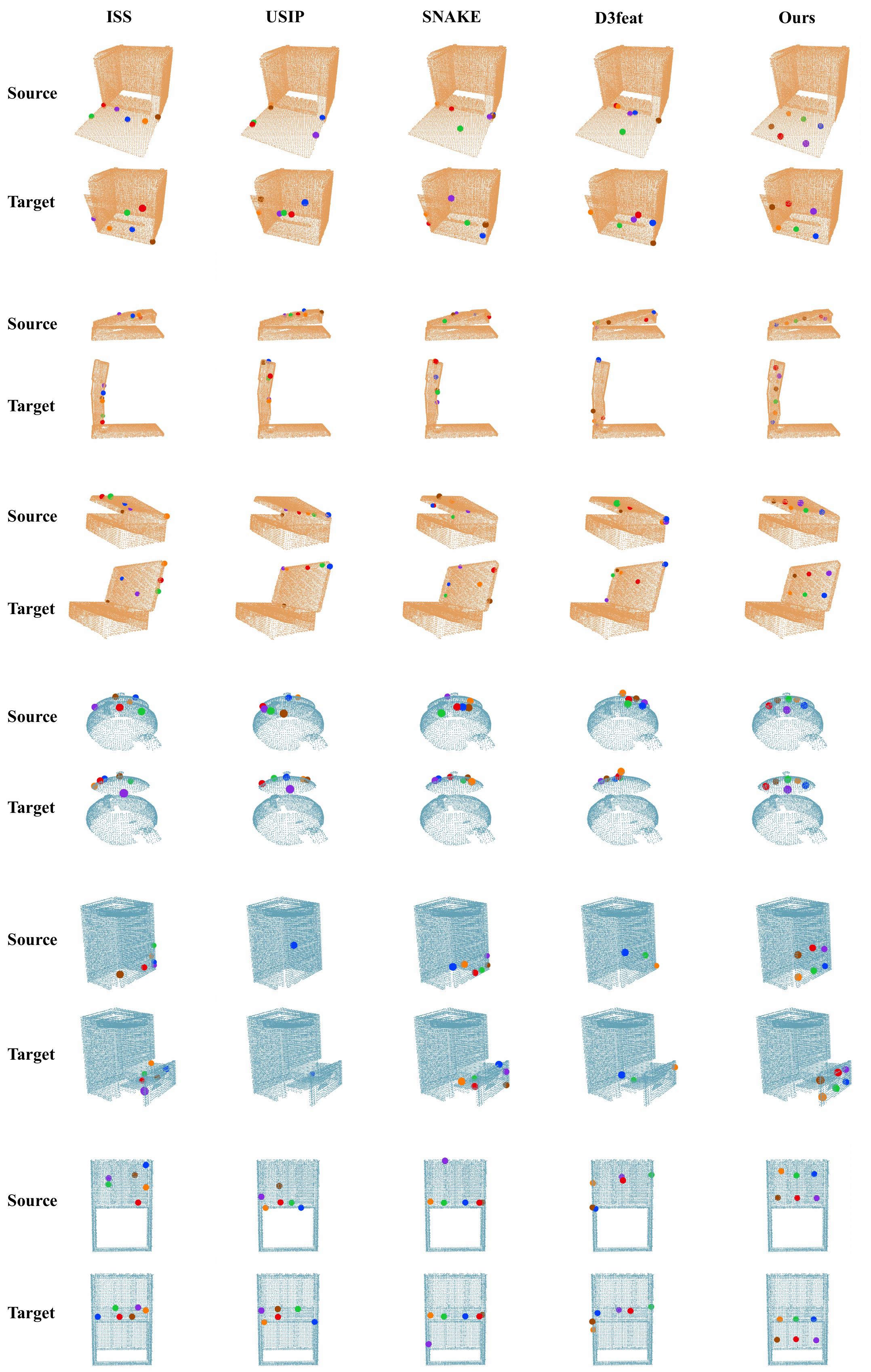}
  \caption{Keypoint temporal consistency comparison for both revolute and prismatic joints.}
  \label{fig:kp_cons}
\end{figure*}

\begin{figure*}[ht]
  \centering
  \includegraphics[width=0.8\linewidth]{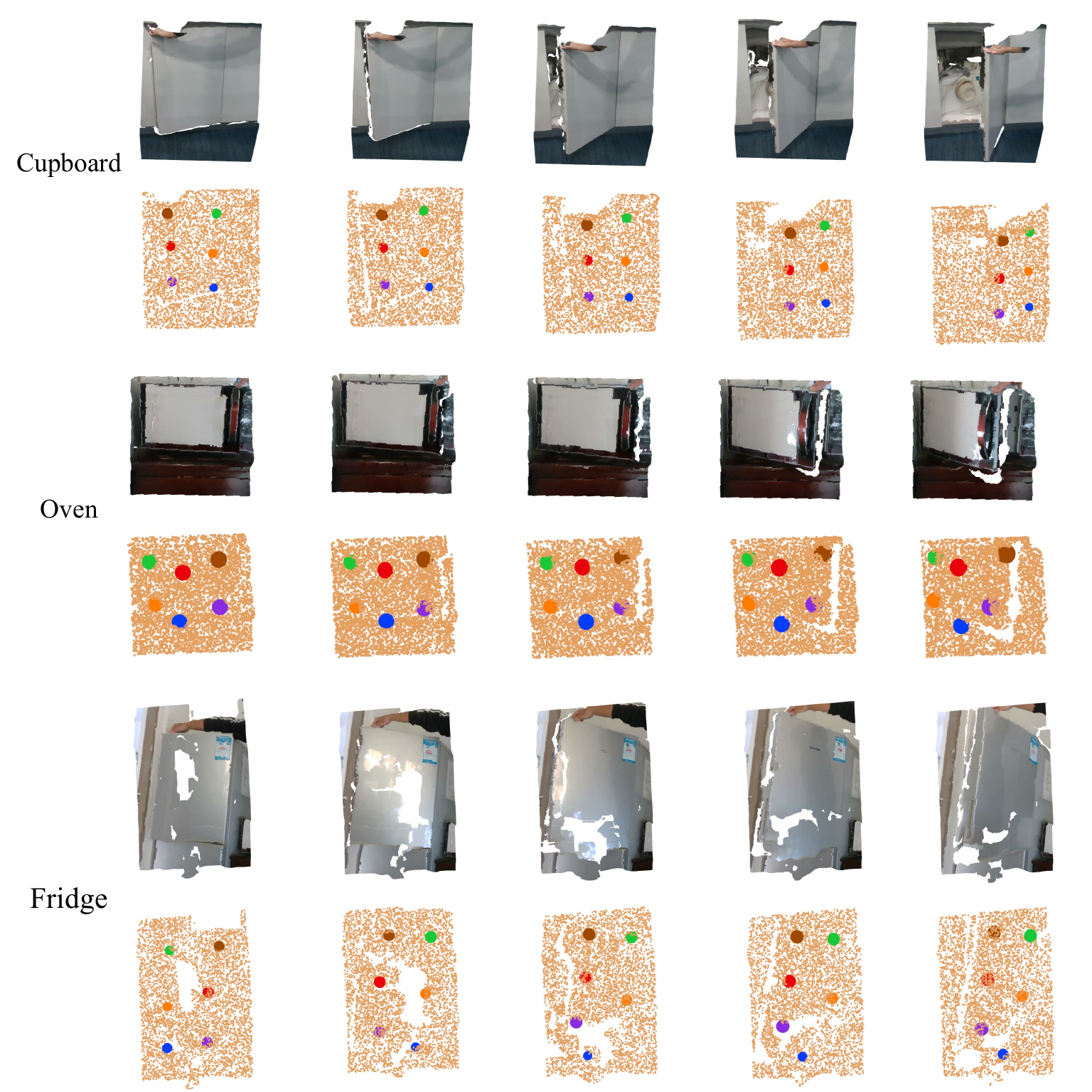}
  \caption{Keypoint consistency of real objects. The input point clouds are cropped by the human labeled bounding box in the first frame of an object video.}
  \label{fig:kp_real}
\end{figure*}

\paragraph{Keypoint Consistency}
We show more qualitative results on keypoint temporal consistency between the same instance with different articulated states in Fig.~\ref{fig:kp_cons}. It can be seen that our method can generate more consistent keypoints than other baselines in both revolute and prismatic joints. 
We also provide visualizations of real objects in Fig.~\ref{fig:kp_real}. Since the real depth image often contains artifacts caused by occlusions, depth discontinuities, or multiple reflections, we adopt the filter method as \cite{wang2019densefusion} used to fill holes in depth image and smooth depth values. Nevertheless, the point clouds may still be incomplete. Despite these artifacts, our method can generally detect spatiotemporally consistent keypoints. We believe the reason is that the implicit geometry decoder can represent the surface occupancy in each continuous input query point, which is robust to the density variation of point clouds.
More intuitive performance can be found in the supplementary video.

\paragraph{Implicit Reconstruction}
Fig. \ref{fig:trans_unseen}-(a) shows the surface reconstruction of the target input, which is based on the transported feature from the source. It demonstrates the effectiveness of the feature transporter and the implicit geometry decoder. Fig. \ref{fig:trans_unseen}-(b) provides more reconstruction results of unseen test categories.

\begin{figure*}[ht]
  \centering
  \includegraphics[width=0.9\linewidth]{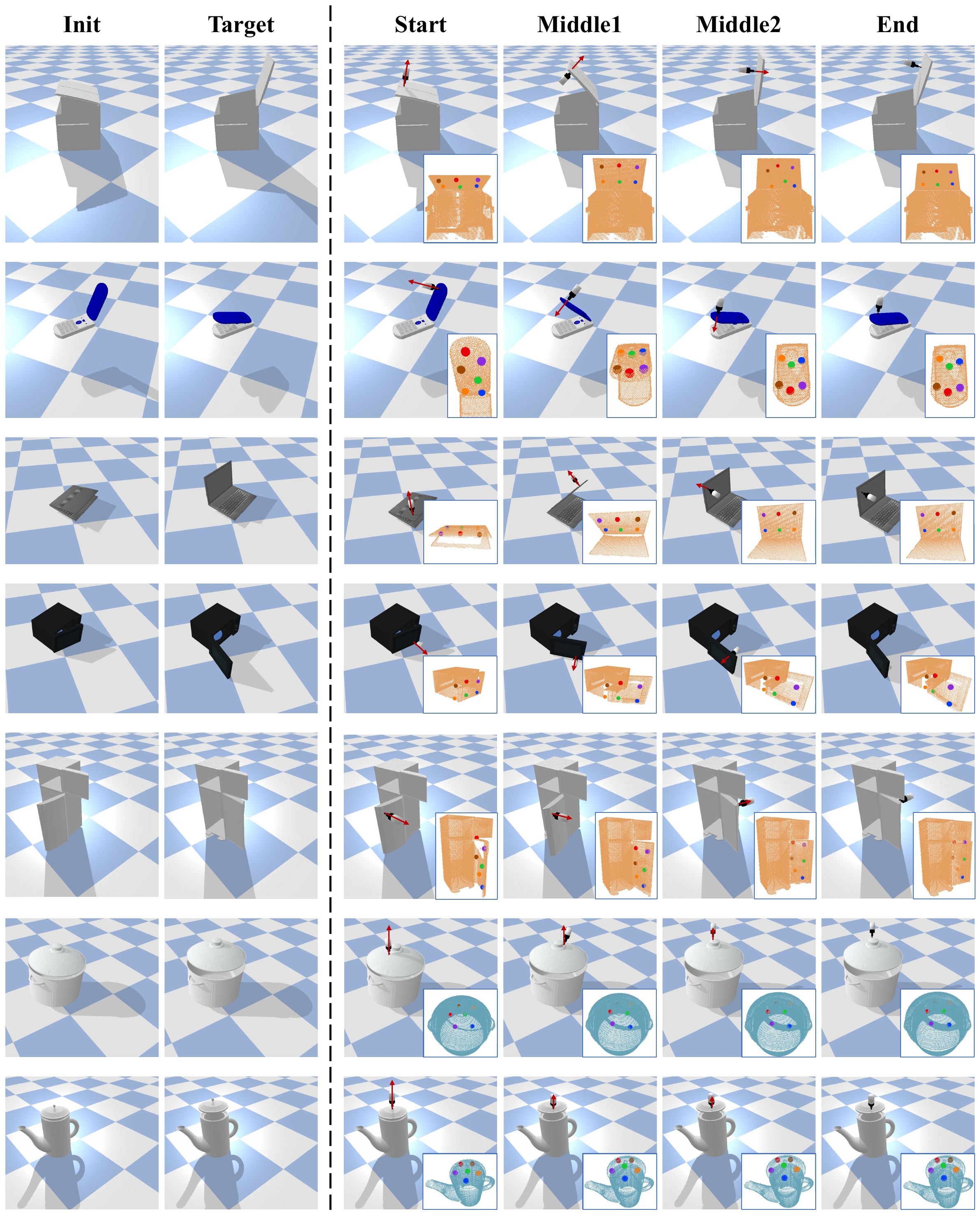}
  \caption{Visualizations of our closed-loop policy for manipulating articulated objects from initial to target states.}
  \label{fig:mani_sim}
\end{figure*}

\begin{figure*}[ht]
  \centering
  \includegraphics[width=0.9
  \linewidth]{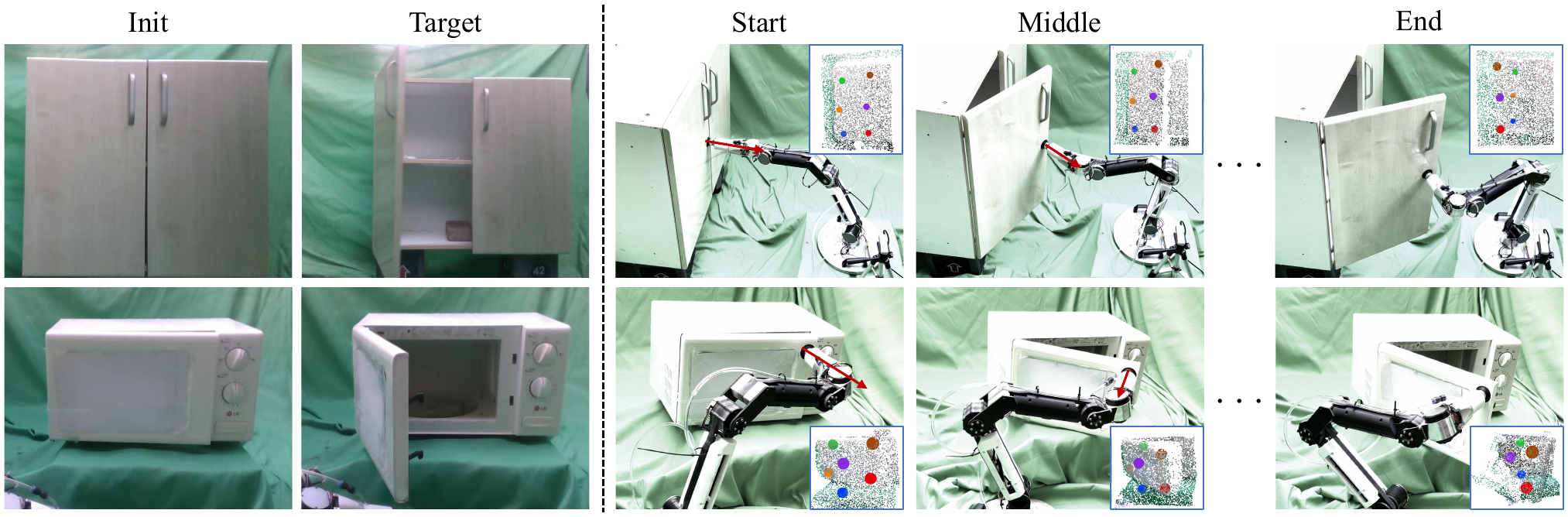}
  \caption{Qualitative results on real object manipulation.}
  \label{fig:mani_real}
\end{figure*}

\paragraph{Goal-conditioned Manipulation}
As shown in Fig.~\ref{fig:mani_sim} (simulation) and Fig.~\ref{fig:mani_real} (real scene), we show visualizations of the closed-loop policy taken to interact with articulated objects from their initial to goal states. 
More qualitative results in simulated and real scenes can be found in the supplementary video.


\section{Failure Cases}
If the number of points of the mobile part is too small, it is difficult for our method to detect accurate keypoints. Moreover, our manipulation strategy fails when the attached keypoint is not on the object's surface, like the point in the drawer.

\end{document}